\documentclass{article}

\usepackage[numbers]{natbib}
\usepackage[final,nonatbib]{neurips_2020}

\usepackage[utf8]{inputenc} %
\usepackage[T1]{fontenc}    %
\usepackage[breaklinks=true,colorlinks,bookmarks=false]{hyperref}
\usepackage{url}            %
\usepackage{booktabs}       %
\usepackage{amsfonts}       %
\usepackage{amsmath}       %
\usepackage{nicefrac}       %
\usepackage{microtype}      %
\usepackage{multirow}
\usepackage{graphicx}
\usepackage{xcolor}         %
\usepackage{enumitem}

\renewcommand{\paragraph}[1]{\noindent {\bf #1}}

\title{Convolutional Generation of Textured 3D Meshes}

\author{%
	Dario Pavllo\\
	Dept.\ of Computer Science\\
	ETH Zurich\\
	\And
	Graham Spinks\\
	Dept.\ of Computer Science\\
	KU Leuven\\
	\AND
	Thomas Hofmann\\
	Dept.\ of Computer Science\\
	ETH Zurich\\
	\And
	Marie-Francine Moens\\
	Dept.\ of Computer Science\\
	KU Leuven\\
	\And
	Aurelien Lucchi\\
	Dept.\ of Computer Science\\
	ETH Zurich\\
}

\begin{document}

\maketitle

\begin{abstract}
While recent generative models for 2D images achieve impressive visual results, they clearly lack the ability to perform 3D reasoning. This heavily restricts the degree of control over generated objects as well as the possible applications of such models.
In this work, we bridge this gap by leveraging recent advances in differentiable rendering. We design a framework that can generate \emph{triangle meshes} and associated high-resolution texture maps, using only 2D supervision from single-view natural images.
A key contribution of our work is the encoding of the mesh and texture as 2D representations, which are semantically aligned and can be easily modeled by a 2D convolutional GAN. We demonstrate the efficacy of our method on Pascal3D+ Cars and CUB, both in an unconditional setting and in settings where the model is conditioned on class labels, attributes, and text. Finally, we propose an evaluation methodology that assesses the mesh and texture quality separately.\looseness=-1

\end{abstract}

\section{Introduction}
\label{sec:introduction}

State-of-the-art image synthesis models based on the GAN framework \cite{goodfellow2014gan} nowadays achieve photorealistic results thanks to a series of key contributions in this area \cite{heusel2017ttur, miyato2018spectral, miyato2018cgans, zhang2018sagan, karras2017progressivegan}. 
A recent trend in this field has been to make generative models more controllable and of better use for downstream applications. This includes works that condition generative models on class labels \cite{miyato2018cgans, zhang2018sagan, brock2018biggan}, text \cite{zhang2017stackgan, zhang2018stackganpp, xu2018attngan, li2019controlgan}, input images \cite{zhu2017cyclegan, isola2017pix2pix}, as well as structured scene layouts such as semantic maps \cite{wang2018pix2pixhd, park2019spade, mo2018instagan}, bounding boxes \cite{zhao2019imagelayout, hong2018inferring, sun2019image}, and scene graphs \cite{johnson2018imagescenegraphs}. While these approaches achieve impressive visual results, they are all based on architectures that fundamentally ignore the concept of image formation. Real-world images depict 2D projections of 3D objects, and explicitly considering this aspect would lead to better generative models that can provide disentangled control over shape, color, pose, lighting, and can better handle spatial phenomena such as occlusions. A recent trend to account for such effects has been to disentangle factors of variation during the generation process in the hope of making it more interpretable \cite{yang2017lrgan, singh2018finegan, karras2017progressivegan, karras2019stylegan}. These approaches potentially learn a hierarchical decomposition of objects, and in some settings (e.g. faces) they can provide some degree of control over pose. However, the pose disentanglement assumptions made by these approaches have been shown to be unrealistic without some form of supervision \cite{locatello2018challenging}, and they have not reached the degree of controllability that a native 3D representation would be capable of. More recent efforts have focused on incorporating 3D information into the model architecture, using either rigid transformations in feature space \cite{nguyen2019hologan} or analysis-by-synthesis \cite{mustikovela2020self}. These approaches represent an interesting middle ground between 2D and 3D generators, although their objective remains 2D image synthesis.\looseness=-1

In this work, we propose a GAN framework for generating \emph{triangle meshes} and associated textures, using only 2D supervision from single-view natural images. In terms of applications, our approach could greatly facilitate content creation for art, movies, video games, virtual reality, as well as augment the possible downstream applications of generative models.
We leverage recent advances in differentiable rendering \cite{loper2014opendr, kato2018n3mr, liu2019softras, chen2019dibr} to incorporate 3D reasoning into our approach. In particular, we initially adopt a reconstruction framework to estimate meshes through a representation we name \emph{convolutional mesh} which consists of a displacement map that deforms a mesh template in its tangent space. This representation is particularly well-suited for 2D convolutional architectures as both the mesh and its texture share the same topology, and the mesh benefits from the spatial smoothness property of convolutions. We then project natural images onto the UV map (mapping between texture coordinates and mesh vertices) and reduce the problem to a 2D modeling task where the representation is independent of the pose of the object. Finally, we train a 2D convolutional GAN in UV space where inputs to the discriminator are masked in order to deal with occlusions.

Our model generates realistic meshes and can easily scale to high-resolution textures (512$\times$512 and possibly more) owing to the precise semantic alignment between maps in UV space, without requiring progressive growing \cite{karras2017progressivegan}. Most importantly, since our model is based exclusively on 2D convolutions, we can easily adapt ideas from state-of-the-art GAN methods for 2D images, and showcase our approach under a wide range of settings: conditional generation from class labels, attributes, text (with and without attention), as well as unconditional generation.
We evaluate our approach on Pascal3D+ Cars \cite{xiang2014pascal} and CUB Birds \cite{wah2011cub}, and propose metrics for evaluating FID scores \cite{heusel2017ttur} on meshes and textures separately as well as collectively. In summary, we make the following contributions:\looseness=-1
\vspace{-1.5mm}
\begin{itemize}[leftmargin=*, itemsep=0pt]
    \item A novel convolutional mesh representation that is smooth by definition, and alongside the texture, is easy to model using standard 2D convolutional GAN architectures.
    \item A GAN framework for producing textured 3D meshes from a pose-independent 2D representation. In particular, in a GAN setting, we are the first to demonstrate full generation of textured triangle meshes using 2D supervision from \emph{natural images}, whereas prior attempts have focused on limited settings supervised on synthetic data without a principled texture learning strategy.
    \item We demonstrate \emph{conditional} generation of 3D meshes from text (with and without an attention mechanism) and show that our model provides disentangled control over shape and appearance.
    \item We release our code and pretrained models at \url{https://github.com/dariopavllo/convmesh}. %
\end{itemize}

\section{Related work}
Deep learning approaches that deal with 3D data typically target either \emph{reconstruction}, where the goal is to predict a 3D mesh from an image, or \emph{generation}, where the goal is to produce meshes from scratch. We review the literature of both tasks as they are relevant to our work.

\paragraph{3D representations.} Early approaches have focused on reconstructing meshes using 3D supervision. These are typically based on voxel grids \cite{girdhar2016learning, choy20163d, zhu2017rethinking, wu2017marrnet, yang20173d, tatarchenko2017octree, hane2017hierarchical}, point clouds \cite{fan2017point}, or signed distance functions \cite{park2019deepsdf}. However, 3D supervision requires ground-truth 3D meshes, which are usually available in synthetic datasets but not for real-world images. Therefore, a related line of research aims at reconstructing meshes using exclusively 2D supervision from images. Similarly, there has been work on voxel representations \cite{yan2016perspective, gwak2017weakly, tulsiani2017multi, BMVC2017_99, tulsiani2018multi, yang2018learning} as well as on point clouds \cite{insafutdinov2018unsupervised}, but these methods require supervision from multiple views which still limits their applicability. More recent approaches lift the requirement of multiple views in order to learn to reconstruct 3D shapes from a single view using a voxel representation \cite{henzler2019escaping}. However, these representations tend to be computationally inefficient and do not explicitly support texture maps.

\paragraph{Differentiable rendering.} Triangle meshes are an established representation in computer graphics, owing to their efficiency as well as flexibility in terms of vertex transformations and texturing. For this reason, they are used in almost every graphics pipeline, ranging from video games to animation. This has motivated a newer line of research where the goal is to predict triangle meshes and texture maps from single images, achieving high-quality visual results \cite{kato2018n3mr, kanazawa2018cmr, chen2019dibr}. The basic building block of these approaches is a \emph{differentiable renderer} (DR), i.e. a renderer that can compute gradients w.r.t. the scene parameters. While early DRs approximate gradients with respect to mesh vertices \cite{loper2014opendr,kato2018n3mr}, newer methods propose fully-differentiable formulations \cite{liu2019softras,chen2019dibr}. Our work is also based on this framework, and specifically we adopt DIB-R \cite{chen2019dibr} because it supports UV mapping.

\paragraph{3D mesh generation.} Analogous to reconstruction methods, 3D object generation has also been demonstrated using voxels \cite{wu2016learning,girdhar2016learning, smith2017improved, xie2018learning, zhu2018visual, balashova2018structure} and point clouds \cite{achlioptas2018learning, gadelha2018multiresolution}, but again, these approaches require some form of 3D supervision which precludes training from natural images, in addition to the texturing limitations highlighted above.
As for triangle meshes, \cite{chen2019dibr} propose a GAN framework where 2D images are discriminated after differentiable rendering, but they rely on multiple views of synthetic objects and cannot directly learn textures from images. Instead, they supervise the generator on textures predicted by a separate model previously trained for reconstruction. This intermediate step results in a noticeable loss of quality, and is absent in our approach, which can learn from natural images directly.
A parallel work to ours \cite{henderson2020leveraging} also leverages 2D data to generate 3D meshes, but they adopt a VAE framework \cite{kingma2013vae} and only predict face colors instead of UV-mapped textures (i.e. \emph{texture maps}), which limits the visual detail of generated objects. An early work \cite{rezende2016unsupervised} generates untextured meshes in a variational framework using reinforcement learning to estimate gradients.
Our work is based on GANs and can explicitly generate high-resolution texture maps which are then mapped to the mesh via UV mapping, enabling an arbitrary level of detail. Unlike \cite{chen2019dibr}, we learn textures directly from natural images, and introduce a pose-independent representation that reduces the problem to a 2D modeling task. Finally, we are not aware of any prior work that can generate 3D meshes from text.\looseness=-1

\section{Method}
\label{sec:method}

\begin{figure}[h]
  \centering
  \includegraphics[width=\textwidth]{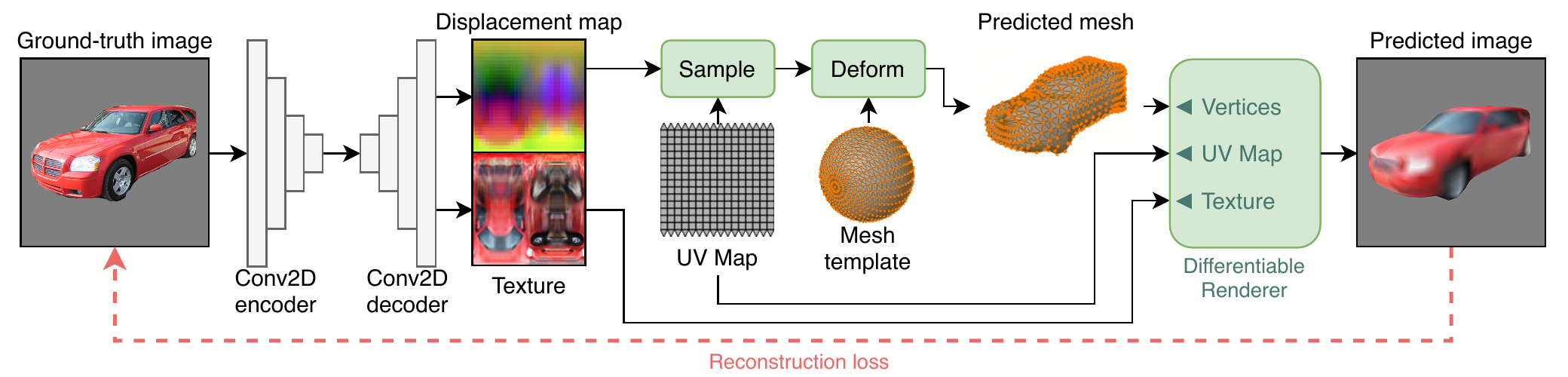}
  \vspace{-6mm}
  \caption{Initial mesh reconstruction using our convolutional mesh representation. This step follows a typical autoencoder setup where the goal is to reconstruct the input image after forcing it through a 3D representation and rendering it. RGB colors in the displacement map correspond to XYZ coordinates.}
  \label{fig:step1}
\end{figure}

\paragraph{Requirements.} Our approach has data requirements similar to recent \emph{reconstruction} methods \cite{kanazawa2018cmr, chen2019dibr}. We require a dataset of single-view natural images, with annotated segmentation masks and pose, or alternatively, keypoints from which the pose can approximately be estimated. If ground-truth masks are not available (as in the ImageNet subset of Pascal3D+), we obtain them from an off-the-shelf segmentation model (we use Mask R-CNN \cite{he2017maskrcnn}), whereas the pose is inferred from the keypoints using structure-from-motion, as was done in \cite{kanazawa2018cmr}. Our approach does not require ground-truth 3D meshes (i.e. 3D supervision) or multiple views of the same object.

\paragraph{Mesh representation.}
As mentioned, we focus on triangle meshes due to their wide adoption in computer graphics, their flexibility in terms of vertex transformations, and their support for texture mapping. Following \cite{kato2018n3mr, kanazawa2018cmr, chen2019dibr}, we use a deformable sphere template with a fixed topology and a static UV map which maps vertices to texture coordinates. Previous work has used fully-connected networks to predict vertex positions, which ignores the topology of the mesh and the spatial correlation between neighboring vertices, essentially treating each vertex as independent. This issue is typically mitigated through regularization, e.g. by combining smoothness \cite{kato2018n3mr} and Laplacian \cite{sorkine2004laplacian} loss penalties. Instead, we propose to regress the mesh through the same deconvolutional network that we use to regress the texture. The output is therefore a \emph{displacement map} (\autoref{fig:step1}), which describes how the mesh should be deformed in its tangent space. Importantly, the displacement map and the texture share the same UV map, which ensures that the maps are topologically aligned (e.g. the vertices corresponding to the beak of a bird are co-located with the color of the beak). This detail is crucial for designing a discriminator that can jointly discriminate mesh and texture, that is, not just the mesh and texture separately, but also how well the texture fits the mesh. Furthermore, our mesh representation is smooth by nature since it benefits from the intrinsic spatial correlation of convolutional layers.
A second major difference in terms of representation is that our mesh template is a \emph{UV sphere} (2-pole sphere as shown in \autoref{fig:step1}), whereas prior work has used ico-spheres. While the latter exhibits a more regular mesh, it cannot be UV-mapped without gaps or arbitrary distortions that make the representation space discontinuous. On the other hand, except for the singularities at the two poles, a UV sphere presents a bijective mapping between vertices and texture coordinates, has a well-defined tangent map, and the circular boundary conditions along the $x$ axis can be neatly incorporated in the model architecture using circular convolutions. Denoting the mesh template as $\mathbf{V}$ (an $N \times 3$ matrix with $N$ vertices described by their $xyz$ coordinates, where each vertex is indexed by $i$), the final position of the $i$-th vertex is computed as $\mathbf{v_i} + \mathbf{R_i} \boldsymbol{\Delta}_\mathbf{v_i}$ where $\boldsymbol{\Delta}_\mathbf{v_i}$ is the output of the model after sampling the displacement map, and $\mathbf{R_i}$ is a precomputed rotation matrix that describes the local normal, tangent, and bitangent of the vertex.%

\begin{figure}[t]
  \centering
  \includegraphics[width=\textwidth]{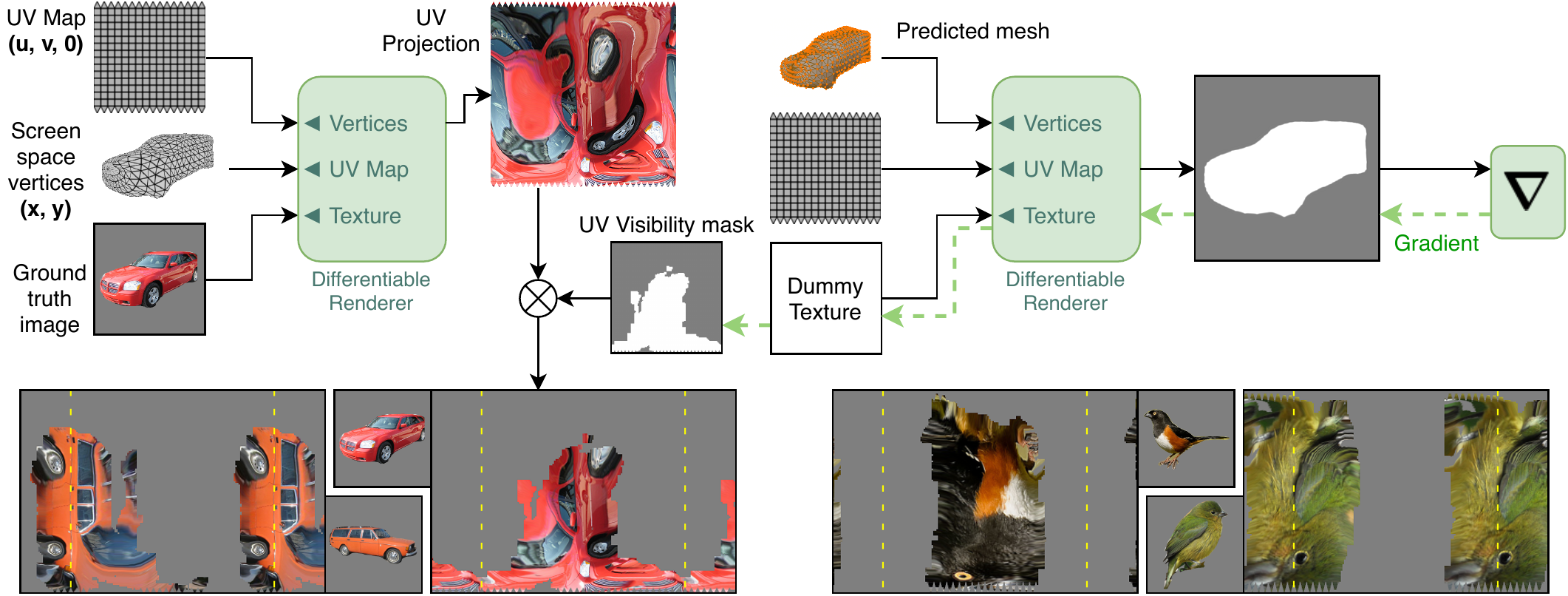}
  \vspace{-6mm}
  \caption{Projection of ground-truth images onto the UV map, producing \emph{pseudo}-ground-truth textures. The bottom row shows additional examples (Pascal3D+ Cars on the left and CUB Birds on the right). The yellow dashed lines represent the boundaries of the textures, which have been extended to highlight the circular boundary conditions along the $x$ axis.}
  \label{fig:step2}
  \vspace{-1mm}
\end{figure}

\subsection{Pose-independent dataset}
\vspace{-1mm}
Our approach initially augments the dataset by estimating a mesh for each training image (\autoref{fig:step1}). The images are then converted into a pose-independent representation (\autoref{fig:step2}), which can be finally modeled by a 2D GAN (\autoref{fig:step3}).

\paragraph{Mesh estimation.} This is a typical reconstruction task where the goal is to reconstruct the mesh from an input image. Our approach is loosely based on \cite{kanazawa2018cmr}, but simplified since we are not interested in performing inference on unseen images. Our formulation can be regarded as a fitting process where we only keep the predicted meshes and discard the model weights/predicted textures. As depicted in \autoref{fig:step1}, the input image is fed to a convolutional encoder, compressed into a vector representation, and decoded through a convolutional decoder which jointly outputs a texture and a displacement map. The predicted texture is only used to facilitate the learning process and produce more semantically-aligned meshes, and is discarded afterwards. The mesh template is deformed as described by the displacement map, and the final result is rendered using a differentiable renderer. The model is trained to minimize the mean squared error (MSE) between the rendered image and the input image. While this generally leads to blurry textures, it does not represent an issue in our case as these textures are discarded. Since we are not interested in performing inference, we do not predict pose or keypoints, nor do we use texture flows or perceptual losses to improve predicted textures. For the camera model, we adopt a weak-perspective model where the pose of an image is described by a rotation $\mathbf{q} \in \mathbb{R}^4$ (a unit quaternion), a scale $s \in \mathbb{R}$, and a screen-space translation $\mathbf{t} \in \mathbb{R}^2$. For Pascal3D+, we augment the projection model with a perspective correction term $z_0$ (further details in the Appendix \ref{sec:appendix-implementation-details}). While these are initially estimated using structure-from-motion on keypoints \cite{kanazawa2018cmr}, we allow the optimizer to fine-tune $s$, $\mathbf{t}$, and $z_0$ (if used), i.e. we additionally optimize with respect to the dataset parameters\footnote{In an inference model this would be detrimental to generalization, but our goal is mesh fitting.}. This leads to a better alignment between rendered masks and ground-truth masks, facilitating the next step. As a side note, we mention that inaccurate camera assumptions (e.g.\ using an orthographic model on photographs that exhibit significant perspective distortion) would most likely not affect the mask alignment or convergence of the model, but might lead to distorted meshes. Nonetheless, our method can work with \emph{any} projection model as long as the camera parameters are known or can be estimated.\looseness=-1

\paragraph{2D discrimination.} The most obvious way to adapt the aforementioned reconstruction framework is to train a GAN where the generator \textbf{G} produces a 3D mesh and the discriminator \textbf{D} discriminates its 2D projection after differentiable rendering, as in \cite{chen2019dibr}. However, we found this strategy to lead to training instabilities due to the discrepancies of the representation being used by \textbf{G} and \textbf{D} (which are respectively pose-independent and pose-dependent). A further complication we observed is an aliasing effect in the gradient from the differentiable renderer. Successful 2D GAN models typically use complementary architectures for \textbf{G} and \textbf{D} (e.g. both convolutional), which motivates our next idea.\looseness=-1

\paragraph{Pose-independent representation.} We instead propose to project ground-truth images onto the UV map of the mesh template, thus reducing the generative model to a 2D GAN that can be trained with existing convolutional techniques. The construction of this representation is depicted in \autoref{fig:step2}, and can be regarded as a form of \emph{inverse rendering}. We treat our previous mesh estimates as if they were texture coordinates, i.e. $(x, y) \rightarrow (u, v)$ ($z$ is dropped), the UV map becomes the mesh to render (a flat surface with $z = 0$), and the texture is the ground-truth image. The result is the projection of the natural image onto the UV map. However, as can be seen in the figure, this process erroneously projects occluded vertices (the back of the car in the example), which should ideally be masked out as visual information associated with them is not available in the 2D image. We therefore mask the projection using a binary \emph{visibility mask}, which describes what parts of the mesh are visible in UV space. The mask is obtained by rendering the mesh using a dummy texture (e.g. all white) and computing its gradient with respect to the texture (we provide implementation details in the Appendix \ref{sec:appendix-implementation-details}). Only \emph{texels} (pixels of the texture) that contribute to the final image (i.e.\ visible ones) will have non-zero gradients, therefore we obtain the visibility mask by thresholding these gradients. The final result is a pose-independent dataset of \emph{pseudo}-ground-truth textures (because they are partially occluded). A useful consequence of this representation is that samples become semantically aligned, i.e.\ the positions of parts such as wheels or eyes are aligned across all images.

\vspace{-1mm}
\subsection{GAN framework}
\vspace{-1mm}

\begin{figure}[h]
  \centering
  \includegraphics[width=\textwidth]{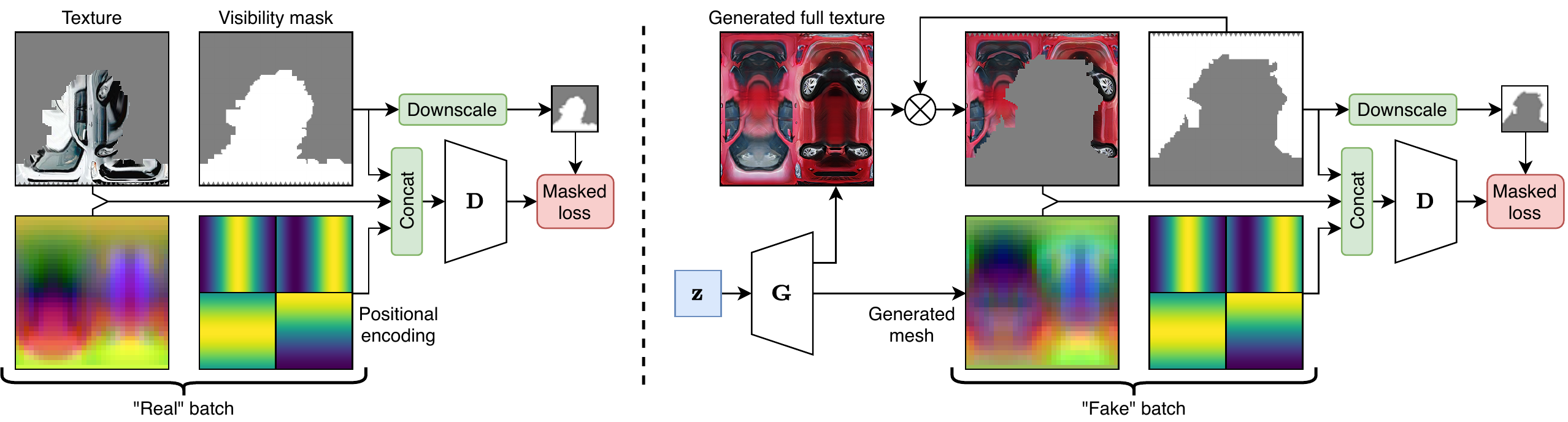}
  \vspace{-6mm}
  \caption{GAN training strategy. \textbf{Left:} discrimination of a ``real'' batch with \emph{pseudo-}ground-truth textures. \textbf{Right:} discrimination of a ``fake'' batch after masking to reflect the ``real'' batch distribution.}
  \label{fig:step3}
\end{figure}

We directly use the estimated displacement maps and the \emph{pseudo}-ground-truth textures to train a convolutional GAN, with the only obstacle that ``real'' textures are masked, while generated textures should ideally be complete. This can be easily dealt with by masking ``fake'' images before they reach \textbf{D}: as shown in \autoref{fig:step3}, we multiply the batch of generated textures with a random sample of visibility masks from the training set. This strategy avoids a distribution mismatch between fake and real textures in \textbf{D}, while acting as a gradient gate such that only gradients from the visible areas will reach the generator. Being convolutional and agnostic to the visibility mask, \textbf{G} will always generate the full texture.\looseness=-1

In terms of architecture, the generator is a convolutional model that outputs both mesh (displacement map) and texture. Mesh and texture can have different resolutions -- in our experiments we use 32$\times$32 for the mesh and up to 512$\times$512 for the texture. To support this, the generator branches out at some point and outputs mesh and texture through two different heads (this is also done in the \emph{mesh estimation} model). The discriminator adopts a multi-scale architecture \cite{wang2018pix2pixhd} (i.e. multiple discriminators trained jointly) and a patch-based loss \cite{isola2017pix2pix} which is masked using the visibility mask scaled to the same resolution as the last feature map. The smallest discriminator discriminates both the mesh and the texture, which is downscaled to the same resolution as the mesh (32$\times$32). It focuses on global aspects of the texture, while discriminating the mesh and how well it fits the texture. The higher-resolution discriminators are only texture discriminators (one for experiments at 256$\times$256, up to two for experiments at 512$\times$512 in which the intermediate one discriminates at 128$\times$128).

\textbf{D} takes as input the displacement map (mesh), the masked texture, the visibility mask, as well as a \emph{soft} positional encoding of the UV map: inspired by attention-based NLP methods that propose a similar idea \cite{vaswani2017transformer} (this is unrelated to our attention method for text conditioning), we add a sinusoidal encoding to the input that gives convolutions a sense of where they are within the image. For a coordinate space $u, v \in [-1, 1]$, we add four channels $\cos(\pi u)$, $\sin(\pi u)$, $\cos(\pi(v/2 + 0.5))$, $\sin(\pi(v/2 + 0.5))$ such that the encoding smoothly wraps around the $u$ (horizontal) axis and is discontinuous along the $v$ (vertical) axis. Giving an absolute sense of position to the model is important as the semantics of the \emph{texels} depend on their absolute position within the UV map, and we show this quantitatively in the ablation study (\autoref{sec:results}).
Finally, the GAN framework allows us to condition the generator on a wide range of inputs: class labels, classes combined with attributes, and text. For the latter, we investigate both an attention mechanism and a method based on a simple fixed-length sentence embedding. We explain how these are implemented in \autoref{sec:implementation-details}.

\section{Experiments}
\vspace{-2mm}
\subsection{Evaluation and datasets}
\label{sec:evaluation}
\vspace{-2mm}
Perceptual metrics such as the Fréchet Inception Distance (FID) \cite{heusel2017ttur} are widely employed for evaluating 2D GANs, as they have been shown to correlate well with human judgment \cite{zhang2018unreasonable}. Although we focus on a  different task, the FID still appears as a natural choice as it can easily be adapted to our task. Therefore, we suggest to evaluate FID scores on rendered 2D projections of generated meshes. To this end, we sample random poses (i.e.\ viewpoints) from the training set as we do not want the evaluation metric to be affected by our choice of poses.
Moreover, this strategy allows us to evaluate mesh and texture separately: in addition to the \emph{Full FID}, we report the \emph{Texture FID}, where we use meshes estimated using the differentiable renderer instead of generated ones, and the \emph{Mesh FID}, where we replace generated textures with \emph{pseudo-}ground-truth ones. In the latter, using real poses ensures that we render the visible part of the \emph{pseudo-}ground-truth texture, and occlusions are minimized. While we mostly rely on the \emph{Full FID} to discuss our results, the individual ones represent a useful tool for analyzing how the model responds to variations of the architecture.  %
Generated samples are rendered at 299$\times$299 (the native resolution of Inception), and ground-truth images are also scaled to this resolution. In the Appendix \ref{sec:appendix-implementation-details}, we provide some visualizations that give more insight into the conceptual differences between these metrics.

We evaluate our method on two datasets with annotated keypoints, and use the implementation of \cite{kanazawa2018cmr} to estimate the pose from keypoints using structure-from-motion.

\paragraph{CUB-200-2011 \cite{wah2011cub}} We use the train/test split of \cite{kanazawa2018cmr}, which consists of $\approx$6k training images and $\approx$5.7k test images. Each image has an annotated class label (out of 200 classes) and 10 captions which we use for text conditioning. Using poses and labels (where applicable) from the training set, we evaluate the FID on test images, although we observe that the FID is almost identical between the two sets.\looseness=-1

\paragraph{Pascal3D+ (P3D) \cite{xiang2014pascal}} We use the \emph{cars} subset, which is the most abundant class in this dataset. Images are part of a low-resolution set (Pascal set) and a newer, high-resolution set from ImageNet \cite{deng2009imagenet}. While we use the same split as \cite{kanazawa2018cmr} to train our mesh estimation model, the GAN is trained only on the ImageNet subset ($\approx$ 4.7k usable images) since we noticed that the images in the Pascal set are too small for practical purposes. We infer segmentation masks using Mask R-CNN \cite{he2017maskrcnn} since they are not available.
The test split of \cite{kanazawa2018cmr} does not contain any ImageNet images, therefore we evaluate FID scores on training images \footnote{Given the already small size of the dataset, we decided not to split it further.}, motivated by our previous observation on CUB.
Finally, to demonstrate conditional generation on this dataset, we collected new annotations for the class (11 shape categories) and color (11 attributes) of each car (details and statistics in the Appendix \ref{sec:appendix-implementation-details}).

\subsection{Implementation details}
\label{sec:implementation-details}

\paragraph{Mesh estimation.} The model (\autoref{fig:step1}) is trained for 1000 epochs using Adam \cite{kingma2014adam}, with an initial learning rate of $10^{-4}$ halved every 250 epochs. We train with a batch size of 50 on a single Pascal GPU, which requires $\approx$12 hours. We use DIB-R \cite{chen2019dibr} for differentiable rendering due to its support for texture mapping and its relatively low overhead.
To stabilize training we adopt a warm-up phase, described in the Appendix \ref{sec:appendix-implementation-details}. In the same section we also describe how we augment the camera model for Pascal3D+. Finally, the detailed architecture of the network can be found in the Appendix \ref{sec:appendix-architecture}.\looseness=-1

\paragraph{GAN architecture.} Since our method is reduced to a 2D generation task, we adopt recent ideas from the 2D convolutional GAN literature. Our generator follows a ResNet architecture where the latent code $\mathbf{z}$ (64D, normally distributed) is injected in the input layer as well as after every convolutional layer through \emph{conditional batch normalization}.  Following \cite{zhang2018sagan, brock2018biggan}, we use spectral normalization \cite{miyato2018spectral} in both \textbf{G} and \textbf{D}, but \textbf{D} does not employ further normalization, e.g. we tried instance normalization but found it detrimental.
We adopt a hinge loss objective (patch-based and masked as described in \autoref{sec:method}), and train for 600 epochs with a constant learning rate of $0.0001$ for \textbf{G} and $0.0004$ for \textbf{D} (two time-scale update rule \cite{heusel2017ttur}). We update \textbf{D} twice per \textbf{G} update, and evaluate the model on a running average of \textbf{G}'s weights ($\beta = 0.999$) as proposed by \cite{yaz2019ganaveraging, karras2017progressivegan, karras2019stylegan, brock2018biggan}.
Detailed aspects about the architecture of our GAN can be found in the Appendix \ref{sec:appendix-architecture}. Training the 512$\times$512 models requires $\approx$ 20 hours on 4 Pascal GPUs, while the 256$\times$256 models require roughly the same time on a single GPU. For all experiments, we use a total batch size of 32 and we employ synchronized batch normalization across multiple GPUs.\looseness=-1

\paragraph{Conditional generation.} In settings conditioned on class labels, we simply concatenate a learnable 64D embedding to $\mathbf{z}$, and use \emph{projection discrimination} \cite{miyato2018cgans} in the last feature map of \textbf{D}. In the P3D experiment with attributes (i.e. colors), we split the embedding into a 32D shape embedding and a 32D color embedding. For text conditioning, we first encode the sentence using the pretrained RNN encoder from \cite{xu2018attngan} (a bidirectional LSTM), and compare \emph{(i)} a simple method where we concatenate the sentence embedding to $\mathbf{z}$ as before, \emph{(ii)} an attention mechanism operating on all hidden states of the RNN. For the latter we add a single attention layer in \textbf{G} right before the mesh/texture branching, operating at 16$\times$16 resolution. Likewise, we modify projection discrimination in \textbf{D} to apply attention on the last feature map. Detailed schemes can be found in the Appendix \ref{sec:appendix-architecture}.

\paragraph{Representation.} Since the UV map of a UV sphere has circular boundary conditions along the horizontal axis, convolutional layers in the discriminator use circular padding horizontally and regular zero-padding vertically. Furthermore, in both the mesh estimation model and the GAN generator, we enforce reflectional symmetry across the $x$ axis as done in \cite{kanazawa2018cmr}, which has the dual benefit of improving quality and halving the computational cost to output a mesh/texture. In this case, convolutions use reflection padding horizontally instead of circular padding. Finally, to deal with the singularities of the UV sphere, the vertex displacements of the north and south pole are respectively taken to be the average of the top and bottom rows of the displacement map.

\subsection{Results}
\label{sec:results}

\begin{table}[!b]
\vspace{-0.5mm}
\caption{\textbf{Left:} FID scores grouped by dataset, texture resolution, and conditioning, both in truncated and untruncated settings. Lower is better; bold = best. \textbf{Right:} Ablation study on CUB with a 512$\times$512 texture resolution. We report truncated FID scores in the truncated setting.}
\label{tbl:main-results}
\vspace{-1.5mm}
\centering
\renewcommand{\arraystretch}{1.15}
\resizebox{0.68\textwidth}{!}{
\begin{tabular}{l|l|l|llll|lll}
                     &                          &              & \multicolumn{4}{c|}{FID (truncated $\sigma$)}                          & \multicolumn{3}{c}{FID (untruncated)}              \\
Dataset              & Tex.\ res.            & Conditioning & $\sigma$ & Full           & Tex.        & Mesh           & Full           & Tex.        & Mesh           \\ \hline\hline
\multirow{4}{*}{CUB} & \multirow{3}{*}{512x512} & None         & 1        & 41.56          & 45.26          & 18.36          & 56.27          & 50.12          & 25.85          \\
                     &                          & Class        & 0.25     & 33.63          & 28.68          & 19.49          & \textbf{41.33} & \textbf{30.60} & 23.28          \\
                     &                          & Text         & 0.5      & \textbf{18.45} & \textbf{22.91} & \textbf{12.05} & 42.66          & 38.95          & \textbf{21.18} \\ \cline{2-10} 
                     & 256x256                  & Class        & 0.25     & 33.55          & 30.92          & 19.39          & 42.61          & 33.31          & 23.37          \\ \hline\hline
\multirow{4}{*}{P3D} & \multirow{3}{*}{512x512} & None         & 1        & 43.09          & 32.70          & 28.62          & 74.74          & 47.99          & 43.23          \\
                     &                          & Class        & 0.75     & \textbf{27.73} & 22.17          & \textbf{23.76} & \textbf{49.56} & \textbf{29.98} & \textbf{34.10} \\
                     &                          & Class+Color  & 0.5      & 31.30          & \textbf{21.70} & 27.75          & 52.55          & 30.29          & 36.32          \\ \cline{2-10} 
                     & 256x256                  & Class+Color  & 0.5      & 39.09          & 26.52          & 36.73          & 63.63          & 36.56          & 46.37         
\end{tabular}
}
\resizebox{0.31\textwidth}{!}{
\begin{tabular}{lll}
                 & FID   & $\Delta$ \\ \hline\hline
Baseline (class) & \textbf{33.63} & 0        \\
No pos.\ encoding     & 43.71 & +10.08   \\
Same G/D updates          & 41.55 & +7.92    \\
InstanceNorm     & 36.38 & +2.75    \\ \hline\hline
Text with attention  & \textbf{18.45} & 0        \\
No attention     & 22.14 & +3.69   
\end{tabular}
}
\vspace{-2.5mm}
\end{table}

\begin{figure}[!b]
  \centering
  \includegraphics[width=\textwidth]{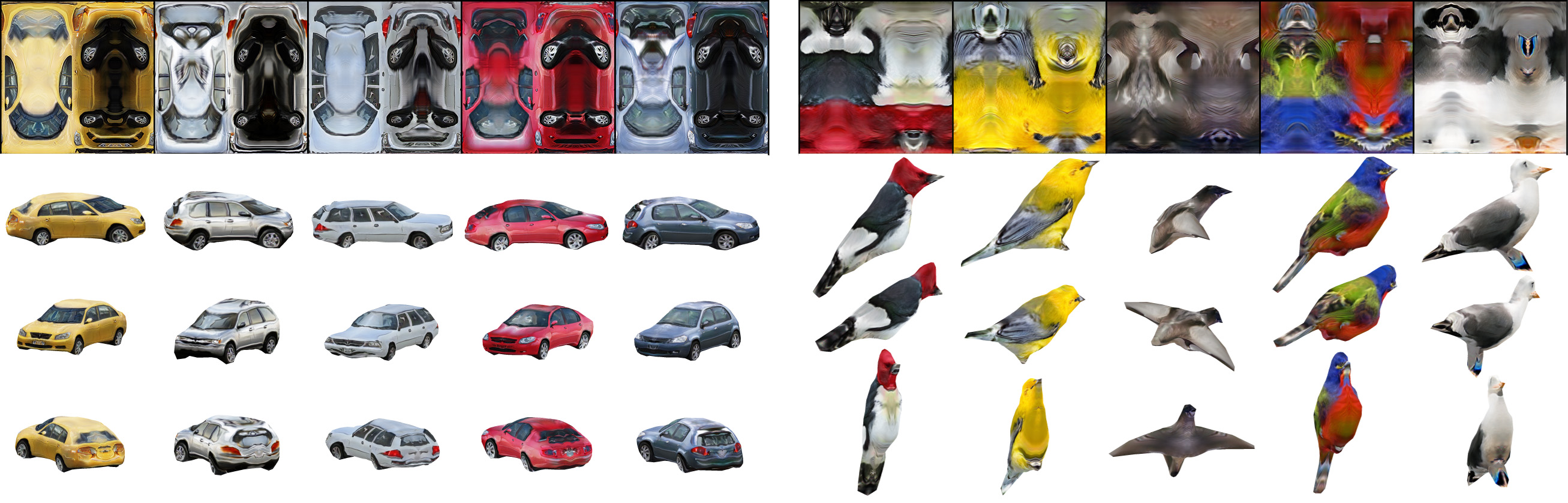}
  \vspace{-7mm}
  \caption{Qualitative results on P3D (left, conditioned on class \emph{and} color) and CUB (right, conditioned on class). Each object is rendered from 3 views, and the top row depicts the unwrapped texture.}
  \label{fig:qualitative}
\end{figure}

\begin{figure}[!b]
  \centering
  \includegraphics[width=\textwidth]{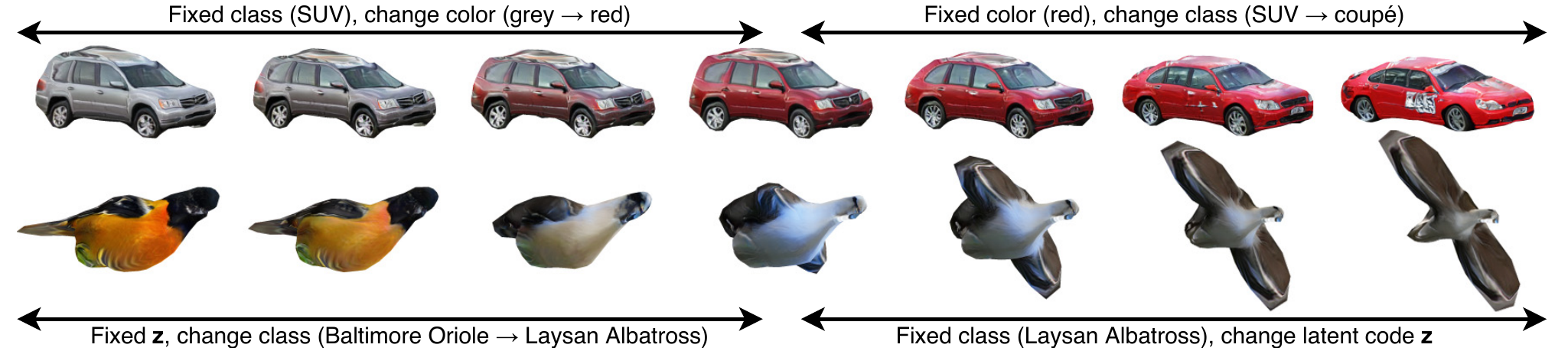}
  \vspace{-6mm}
  \caption{Interpolation over conditioning inputs, which highlights that our model learns a structured latent space where factors of variation of both shape and texture are relatively disentangled.}
  \label{fig:interpolation}
\end{figure}

\begin{figure}[!b]
  \centering
  \includegraphics[width=\textwidth]{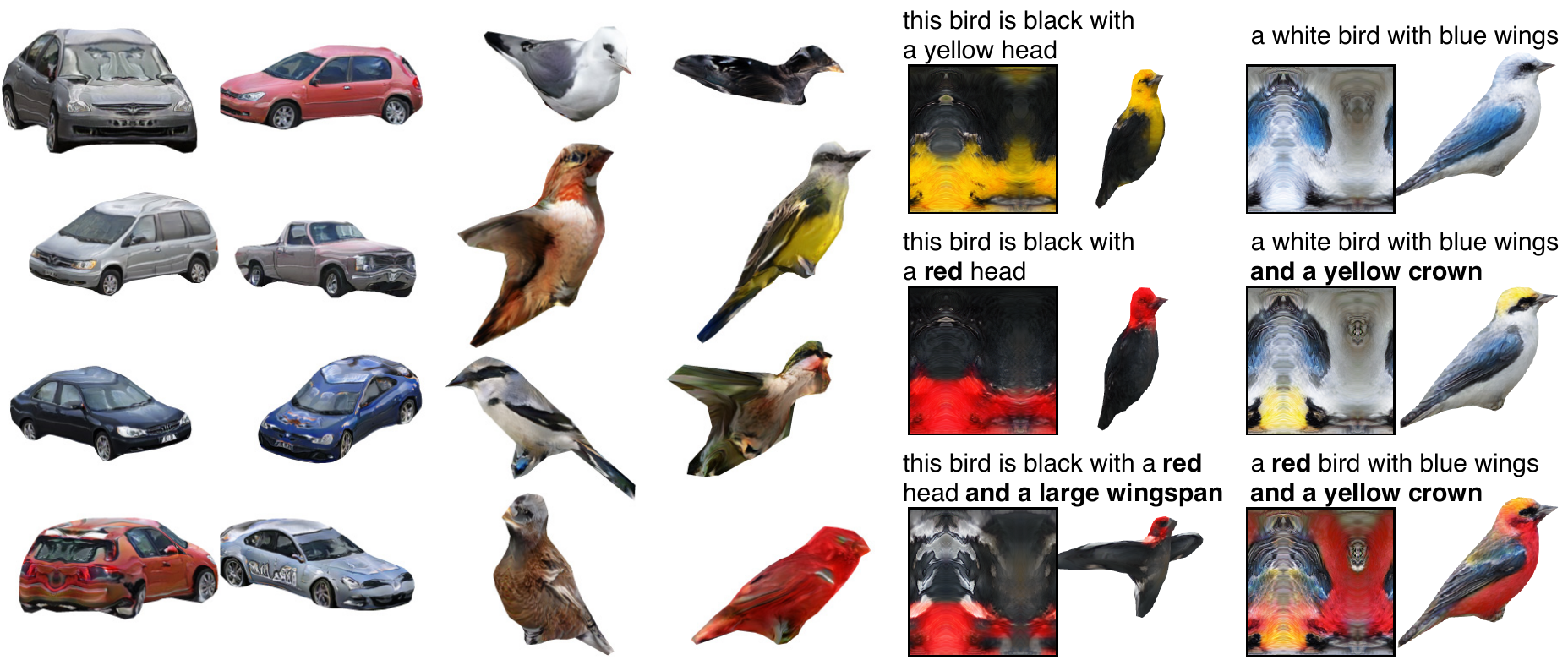}
  \vspace{-6mm}
  \caption{\textbf{Left:} generated meshes rendered from random views on P3D and CUB, both conditioned on a random set of classes. \textbf{Right:} generation from text on CUB, which allows for fine control over both texture and mesh. We modify captions incrementally, where changes are highlighted in bold.}
  \label{fig:random-text-results}
  \vspace{-3mm}
\end{figure}

\paragraph{Quantitative results.} We report our main results in \autoref{tbl:main-results} (left). For CUB, we compare settings where the model is conditioned on class labels, captions (using the attention model), and no conditioning at all. 
For P3D, we compare unconditional generation and conditional generation on class labels (i.e. car shapes) as well as classes \emph{plus} colors. We evaluate the FID every 20 epochs and report the model with the best \emph{Full FID} in the table. Since there is no prior work to which we can compare under our setting, we set baselines on these two datasets. As proposed by \cite{brock2018biggan}, we found it useful to sample latent codes $\mathbf{z}$ from a truncated Gaussian (only at inference time), which trades off sample diversity for quality and considerably improves FID scores. For each setting we specify the optimal truncation $\sigma$, but we also report scores in an untruncated setting as these are more directly comparable. As expected, conditional GANs result in better scores than their unconditional counterparts (with the text model being the best), but we generally observe that our approach is stable under all settings.

\paragraph{Ablation study.} We conduct an ablation study in \autoref{tbl:main-results} (right). The results in the top section are relative to the 512$\times$512 CUB model conditioned on classes (truncated FID). Removing the positional encoding from the discriminator leads to a significant FID degradation (+10.08), suggesting that giving convolutions a sense of absolute position in UV space is an important aspect of our approach. Likewise, updating \textbf{D} as often as \textbf{G} has a significant negative impact (+7.92) compared to two \textbf{D} updates per \textbf{G} update. Using instance normalization in \textbf{D} also leads to a slight degradation (+2.75), but beyond that we observe that, while training appears to converge faster initially, it rapidly becomes unstable. In the bottom section of the table, we compare the text attention model (baseline) to a model where a fixed-length sentence vector is simply concatenated to $\mathbf{z}$ (as in the other conditional models). The results show that the model effectively exploits the attention mechanism with the added benefit of being more interpretable.

\paragraph{Qualitative results.} \autoref{fig:qualitative} shows a few generated meshes rendered from multiple views, as well as the corresponding textures. While results on CUB are generally of high visual quality, we observe that the back of the cars in P3D present some artifacts. After further investigation, we found that the dataset is very imbalanced, with only 10--20\% of the images depicting the back of the car and the majority depicting the front. Therefore, this issue could in principle be mitigated with more training data. In \autoref{fig:interpolation} we show that the latent space of our models is structured. We interpolate over different factors of variation using spherical interpolation and observe that they are relatively disentangled, enabling isolated control over shape, color, and style in addition to the pose disentanglement guaranteed by the 3D representation itself. \autoref{fig:random-text-results} shows results rendered from random viewpoints (the scenario on which we evaluate FID scores) as well as generation conditioned on text, which enables precise control over both shape and appearance. Finally, in the Appendix \ref{sec:appendix-results} we show a wider range of qualitative results.\looseness=-1

\paragraph{2D GAN baseline.} An interesting baseline for generating 3D meshes is to first train a 2D GAN using a state-of-the-art architecture (e.g. StyleGAN \cite{karras2019stylegan}), and then run a 3D mesh reconstruction model on top of the generated 2D images. First, we note that such a baseline would not exhibit the properties of a true 3D representation, such as pose disentangled from shape. Additionally, the reconstruction model would have difficulties dealing with occlusions, since it can only reliably infer information visible in the 2D image. To substantiate our observation, we investigate this baseline empirically: we train the 3D reconstruction model of \cite{kanazawa2018cmr} for 1000 epochs on CUB training images with an empty background (our setting). Evaluating this model on \emph{training images} achieves an FID of 85.8 on reconstructions rendered from ground-truth viewpoints, which is already worse than all of our baselines and establishes a lower bound. If we run the model on CUB images produced by StyleGAN \cite{karras2019stylegan} and evaluate the FID on renderings from sampled viewpoints, the FID further degrades to 101.9.

\begin{figure}[h]
  \centering
  \includegraphics[width=\textwidth]{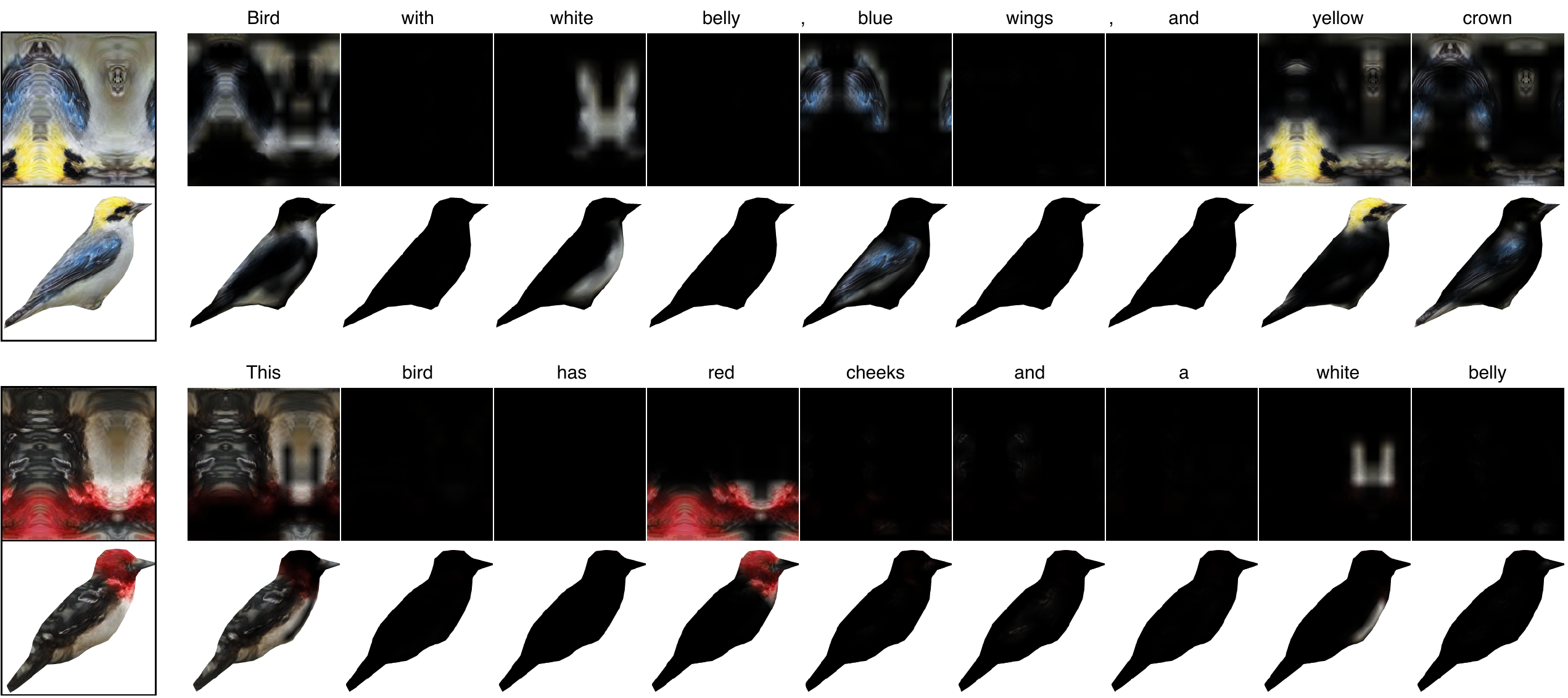}
  \vspace{-5mm}
  \caption{Visualization of the attention mechanism on our CUB model conditioned on text. The attention maps are visualized in UV space (first row) as well as on the rendered mesh (second row). In this particular bidirectional LSTM model, active tokens typically correspond to the adjectives that precede body parts. The first and last tokens are also active because the sentence representation does not comprise explicit sentence delimiters. %
  }
  \label{fig:attention}
  \vspace{-2mm}
\end{figure}

\paragraph{Attention mechanism.} Similar to other attention-based GANs conditioned on text \cite{xu2018attngan}, our attention mechanism can be easily visualized. Interestingly, since the attention is applied to our pose-independent representation in UV space and not on flat 2D images, our attention maps can be visualized both in UV space and on 2D renderings, as we show in \autoref{fig:attention}. Furthermore, our process is more interpretable and semantically meaningful. For instance, prompts that refer to a specific part of the object (e.g. ``yellow crown'', ``red cheeks'') activate the same area within the UV map. Most importantly, these correspondences are learned in an unsupervised fashion and are aligned among different images owing to our pose-independent representation.

\vspace{-1mm}
\section{Conclusion}
\vspace{-2mm}
We propose a GAN-based framework for generating 3D meshes. Our approach can generate triangle meshes and corresponding texture maps at high resolution (512$\times$512 and possibly more), requiring only 2D supervision from single-view natural images. We evaluate our method on Pascal3D+ Cars and CUB Birds, and showcase it under a wide range of conditional settings to demonstrate its high level of adaptation.
Nonetheless, we have only scratched the surface of what can be done with this framework. Our approach can be enriched by employing different forms of supervision (e.g.\ semi-supervision by combining 3D supervision from synthetic datasets with 2D supervision from natural images) as well as incorporating more conditional information that would allow the model to disentangle further aspects of variation (e.g.\ lighting). In the future, we would also like to experiment with larger datasets, and apply the approach to full-scene generation. A viable option is to decompose background and foreground generation as in \cite{pavllo2020stylesemantics}, and use a 3D mesh generator for foreground objects.\looseness=-1

\clearpage
\newpage
\section*{Broader Impact}

Our line of research can positively benefit the video game and film industries, both of which impact the life of billions of users. The ability to partially automate the construction of tailored 3D shapes with textures has the potential to reduce costs and timelines by lessening tedious work. The impact on jobs in this area is likely to be minimal as this work is usually performed by specialists whose expertise could be redirected to more creative tasks \cite{stateofai}. Other areas like education and arts could benefit from the ability to bring new concepts to life in an (interactive) 3D environment. Additionally, mesh generation is a hard problem that is likely to be central in many research areas and industry applications going forward.

Adversely, generative models can be used toward fake content creation. The negative societal impact of our method on image generation is likely small as many image modification tools have existed for years \cite{izitru}. In the longer term, approaches that involve 3D generation might facilitate manipulation of fake video sequences which is harder to achieve with modern software tools. Such applications bring on concerns over exploitation, privacy, political manipulation and the undermining of public institutions. Comparable considerations have already become part of public discourse. A range of approaches have been suggested to tackle such issues ranging from technological, legal, and market solutions. For a more in-depth overview of this discussion we refer to \cite{chesney2019deep}.

\begin{ack}
This work was partly supported by the Swiss National Science Foundation (SNF), grant \#176004, and Research Foundation Flanders (FWO), grant G078618N. Additionally, Graham Spinks and Marie-Francine Moens are partly funded by an ERC Advanced Grant, \#788506.
\end{ack}

{
\small
\bibliographystyle{abbrv}
\bibliography{main}

\begin{thebibliography}{10}

\bibitem{achlioptas2018learning}
P.~Achlioptas, O.~Diamanti, I.~Mitliagkas, and L.~Guibas.
\newblock Learning representations and generative models for 3d point clouds.
\newblock In {\em International Conference on Machine Learning}, pages 40--49,
  2018.

\bibitem{stateofai}
State of {AI} in animation.
\newblock \url{https://theblog.adobe.com/state-of-ai-in-animation/}.
\newblock Accessed: 2020-05-31.

\bibitem{balashova2018structure}
E.~Balashova, V.~Singh, J.~Wang, B.~Teixeira, T.~Chen, and T.~Funkhouser.
\newblock Structure-aware shape synthesis.
\newblock In {\em 2018 International Conference on 3D Vision (3DV)}, pages
  140--149. IEEE, 2018.

\bibitem{brock2018biggan}
A.~Brock, J.~Donahue, and K.~Simonyan.
\newblock Large scale {GAN} training for high fidelity natural image synthesis.
\newblock In {\em International Conference on Learning Representations (ICLR)},
  2019.

\bibitem{chen2019dibr}
W.~Chen, H.~Ling, J.~Gao, E.~Smith, J.~Lehtinen, A.~Jacobson, and S.~Fidler.
\newblock Learning to predict 3d objects with an interpolation-based
  differentiable renderer.
\newblock In {\em Neural Information Processing Systems}, pages 9605--9616,
  2019.

\bibitem{chesney2019deep}
B.~Chesney and D.~Citron.
\newblock Deep fakes: A looming challenge for privacy, democracy, and national
  security.
\newblock {\em Calif. L. Rev.}, 107:1753, 2019.

\bibitem{choy20163d}
C.~B. Choy, D.~Xu, J.~Gwak, K.~Chen, and S.~Savarese.
\newblock 3d-r2n2: A unified approach for single and multi-view 3d object
  reconstruction.
\newblock In {\em European conference on computer vision}, pages 628--644.
  Springer, 2016.

\bibitem{deng2009imagenet}
J.~Deng, W.~Dong, R.~Socher, L.-J. Li, K.~Li, and L.~Fei-Fei.
\newblock Imagenet: A large-scale hierarchical image database.
\newblock In {\em IEEE Conference on Computer Vision and Pattern Recognition
  (CVPR)}, pages 248--255. IEEE, 2009.

\bibitem{fan2017point}
H.~Fan, H.~Su, and L.~J. Guibas.
\newblock A point set generation network for 3d object reconstruction from a
  single image.
\newblock In {\em Proceedings of the IEEE Conference on Computer Vision and
  Pattern Recognition}, pages 605--613, 2017.

\bibitem{izitru}
Photo tampering throughout history.
\newblock \url{http://pth.izitru.com/}.
\newblock Accessed: 2020-05-31.

\bibitem{gadelha2018multiresolution}
M.~Gadelha, R.~Wang, and S.~Maji.
\newblock Multiresolution tree networks for 3d point cloud processing.
\newblock In {\em Proceedings of the European Conference on Computer Vision
  (ECCV)}, pages 103--118, 2018.

\bibitem{girdhar2016learning}
R.~Girdhar, D.~F. Fouhey, M.~Rodriguez, and A.~Gupta.
\newblock Learning a predictable and generative vector representation for
  objects.
\newblock In {\em European Conference on Computer Vision}, pages 484--499.
  Springer, 2016.

\bibitem{goodfellow2014gan}
I.~Goodfellow, J.~Pouget-Abadie, M.~Mirza, B.~Xu, D.~Warde-Farley, S.~Ozair,
  A.~Courville, and Y.~Bengio.
\newblock Generative adversarial nets.
\newblock In {\em Neural Information Processing Systems}, pages 2672--2680,
  2014.

\bibitem{gwak2017weakly}
J.~Gwak, C.~B. Choy, M.~Chandraker, A.~Garg, and S.~Savarese.
\newblock Weakly supervised 3d reconstruction with adversarial constraint.
\newblock In {\em 2017 International Conference on 3D Vision (3DV)}, pages
  263--272. IEEE, 2017.

\bibitem{hane2017hierarchical}
C.~H{\"a}ne, S.~Tulsiani, and J.~Malik.
\newblock Hierarchical surface prediction for 3d object reconstruction.
\newblock In {\em 2017 International Conference on 3D Vision (3DV)}, pages
  412--420. IEEE, 2017.

\bibitem{he2017maskrcnn}
K.~He, G.~Gkioxari, P.~Doll{\'a}r, and R.~Girshick.
\newblock Mask {R-CNN}.
\newblock In {\em IEEE International Conference on Computer Vision (ICCV)},
  pages 2961--2969, 2017.

\bibitem{henderson2020leveraging}
P.~Henderson, V.~Tsiminaki, and C.~H. Lampert.
\newblock Leveraging 2d data to learn textured 3d mesh generation.
\newblock In {\em Proceedings of the IEEE Conference on Computer Vision and
  Pattern Recognition}, 2020.

\bibitem{henzler2019escaping}
P.~Henzler, N.~J. Mitra, and T.~Ritschel.
\newblock Escaping plato's cave: 3d shape from adversarial rendering.
\newblock In {\em Proceedings of the IEEE International Conference on Computer
  Vision}, pages 9984--9993, 2019.

\bibitem{heusel2017ttur}
M.~Heusel, H.~Ramsauer, T.~Unterthiner, B.~Nessler, and S.~Hochreiter.
\newblock {GANs} trained by a two time-scale update rule converge to a local
  {N}ash equilibrium.
\newblock In {\em Neural Information Processing Systems}, pages 6626--6637,
  2017.

\bibitem{hong2018inferring}
S.~Hong, D.~Yang, J.~Choi, and H.~Lee.
\newblock Inferring semantic layout for hierarchical text-to-image synthesis.
\newblock In {\em IEEE Conference on Computer Vision and Pattern Recognition
  (CVPR)}, pages 7986--7994, 2018.

\bibitem{insafutdinov2018unsupervised}
E.~Insafutdinov and A.~Dosovitskiy.
\newblock Unsupervised learning of shape and pose with differentiable point
  clouds.
\newblock In {\em Advances in Neural Information Processing Systems}, pages
  2802--2812, 2018.

\bibitem{isola2017pix2pix}
P.~Isola, J.-Y. Zhu, T.~Zhou, and A.~A. Efros.
\newblock Image-to-image translation with conditional adversarial networks.
\newblock In {\em IEEE Conference on Computer Vision and Pattern Recognition
  (CVPR)}, pages 1125--1134, 2017.

\bibitem{johnson2018imagescenegraphs}
J.~Johnson, A.~Gupta, and L.~Fei-Fei.
\newblock Image generation from scene graphs.
\newblock In {\em IEEE Conference on Computer Vision and Pattern Recognition
  (CVPR)}, pages 1219--1228, 2018.

\bibitem{kanazawa2018cmr}
A.~Kanazawa, S.~Tulsiani, A.~A. Efros, and J.~Malik.
\newblock Learning category-specific mesh reconstruction from image
  collections.
\newblock In {\em European Conference on Computer Vision (ECCV)}, 2018.

\bibitem{karras2017progressivegan}
T.~Karras, T.~Aila, S.~Laine, and J.~Lehtinen.
\newblock Progressive growing of {GAN}s for improved quality, stability, and
  variation.
\newblock In {\em International Conference on Learning Representations (ICLR)},
  2018.

\bibitem{karras2019stylegan}
T.~Karras, S.~Laine, and T.~Aila.
\newblock A style-based generator architecture for generative adversarial
  networks.
\newblock In {\em IEEE Conference on Computer Vision and Pattern Recognition
  (CVPR)}, pages 4401--4410, 2019.

\bibitem{kato2018n3mr}
H.~Kato, Y.~Ushiku, and T.~Harada.
\newblock Neural {3D} mesh renderer.
\newblock In {\em IEEE Conference on Computer Vision and Pattern Recognition
  (CVPR)}, pages 3907--3916, 2018.

\bibitem{kingma2014adam}
D.~P. Kingma and J.~Ba.
\newblock Adam: A method for stochastic optimization.
\newblock In {\em International Conference on Learning Representions (ICLR)},
  2014.

\bibitem{kingma2013vae}
D.~P. Kingma and M.~Welling.
\newblock Auto-encoding variational {B}ayes.
\newblock In {\em International Conference on Learning Representations (ICLR)},
  2014.

\bibitem{li2019controlgan}
B.~Li, X.~Qi, T.~Lukasiewicz, and P.~H.~S. Torr.
\newblock Controllable text-to-image generation.
\newblock In {\em Neural Information Processing Systems}, December 2019.

\bibitem{liu2019softras}
S.~Liu, T.~Li, W.~Chen, and H.~Li.
\newblock Soft rasterizer: A differentiable renderer for image-based 3d
  reasoning.
\newblock In {\em IEEE International Conference on Computer Vision (ICCV)},
  pages 7708--7717, 2019.

\bibitem{locatello2018challenging}
F.~Locatello, S.~Bauer, M.~Lucic, S.~Gelly, B.~Sch{\"o}lkopf, and O.~Bachem.
\newblock Challenging common assumptions in the unsupervised learning of
  disentangled representations.
\newblock In {\em International Conference on Machine Learning (ICML)}, 2019.

\bibitem{loper2014opendr}
M.~M. Loper and M.~J. Black.
\newblock Opendr: An approximate differentiable renderer.
\newblock In {\em European Conference on Computer Vision (ECCV)}, pages
  154--169. Springer, 2014.

\bibitem{miyato2018spectral}
T.~Miyato, T.~Kataoka, M.~Koyama, and Y.~Yoshida.
\newblock Spectral normalization for generative adversarial networks.
\newblock In {\em International Conference on Learning Representations (ICLR)},
  2018.

\bibitem{miyato2018cgans}
T.~Miyato and M.~Koyama.
\newblock c{GAN}s with projection discriminator.
\newblock In {\em International Conference on Learning Representations (ICLR)},
  2018.

\bibitem{mo2018instagan}
S.~Mo, M.~Cho, and J.~Shin.
\newblock Instance-aware image-to-image translation.
\newblock In {\em International Conference on Learning Representations (ICLR)},
  2019.

\bibitem{mustikovela2020self}
S.~K. Mustikovela, V.~Jampani, S.~D. Mello, S.~Liu, U.~Iqbal, C.~Rother, and
  J.~Kautz.
\newblock Self-supervised viewpoint learning from image collections.
\newblock In {\em IEEE Conference on Computer Vision and Pattern Recognition
  (CVPR)}, pages 3971--3981, 2020.

\bibitem{nguyen2019hologan}
T.~Nguyen-Phuoc, C.~Li, L.~Theis, C.~Richardt, and Y.-L. Yang.
\newblock Hologan: Unsupervised learning of 3d representations from natural
  images.
\newblock In {\em IEEE International Conference on Computer Vision (ICCV)},
  pages 7588--7597, 2019.

\bibitem{odena2016deconvolution}
A.~Odena, V.~Dumoulin, and C.~Olah.
\newblock Deconvolution and checkerboard artifacts.
\newblock {\em Distill}, 2016.

\bibitem{park2019deepsdf}
J.~J. Park, P.~Florence, J.~Straub, R.~Newcombe, and S.~Lovegrove.
\newblock {DeepSDF}: Learning continuous signed distance functions for shape
  representation.
\newblock In {\em IEEE Conference on Computer Vision and Pattern Recognition
  (CVPR)}, pages 165--174, 2019.

\bibitem{park2019spade}
T.~Park, M.-Y. Liu, T.-C. Wang, and J.-Y. Zhu.
\newblock Semantic image synthesis with spatially-adaptive normalization.
\newblock In {\em IEEE Conference on Computer Vision and Pattern Recognition
  (CVPR)}, pages 2337--2346, 2019.

\bibitem{pavllo2020stylesemantics}
D.~Pavllo, A.~Lucchi, and T.~Hofmann.
\newblock Controlling style and semantics in weakly-supervised image
  generation.
\newblock In {\em European Conference on Computer Vision (ECCV)}, 2020.

\bibitem{rezende2016unsupervised}
D.~J. Rezende, S.~A. Eslami, S.~Mohamed, P.~Battaglia, M.~Jaderberg, and
  N.~Heess.
\newblock Unsupervised learning of 3d structure from images.
\newblock In {\em Advances in Neural Information Processing Systems}, pages
  4996--5004, 2016.

\bibitem{singh2018finegan}
K.~K. Singh, U.~Ojha, and Y.~J. Lee.
\newblock {FineGAN}: Unsupervised hierarchical disentanglement for fine-grained
  object generation and discovery.
\newblock In {\em IEEE Conference on Computer Vision and Pattern Recognition
  (CVPR)}, 2019.

\bibitem{smith2017improved}
E.~J. Smith and D.~Meger.
\newblock Improved adversarial systems for 3d object generation and
  reconstruction.
\newblock In {\em Conference on Robot Learning}, pages 87--96, 2017.

\bibitem{sorkine2004laplacian}
O.~Sorkine, D.~Cohen-Or, Y.~Lipman, M.~Alexa, C.~R{\"o}ssl, and H.-P. Seidel.
\newblock Laplacian surface editing.
\newblock In {\em Eurographics/ACM SIGGRAPH Symposium on Geometry Processing},
  pages 175--184, 2004.

\bibitem{sun2019image}
W.~Sun and T.~Wu.
\newblock Image synthesis from reconfigurable layout and style.
\newblock In {\em IEEE International Conference on Computer Vision (ICCV)},
  pages 10531--10540, 2019.

\bibitem{tatarchenko2017octree}
M.~Tatarchenko, A.~Dosovitskiy, and T.~Brox.
\newblock Octree generating networks: Efficient convolutional architectures for
  high-resolution 3d outputs.
\newblock In {\em Proceedings of the IEEE International Conference on Computer
  Vision}, pages 2088--2096, 2017.

\bibitem{tulsiani2018multi}
S.~Tulsiani, A.~A. Efros, and J.~Malik.
\newblock Multi-view consistency as supervisory signal for learning shape and
  pose prediction.
\newblock In {\em Proceedings of the IEEE Conference on Computer Vision and
  Pattern Recognition}, pages 2897--2905, 2018.

\bibitem{tulsiani2017multi}
S.~Tulsiani, T.~Zhou, A.~A. Efros, and J.~Malik.
\newblock Multi-view supervision for single-view reconstruction via
  differentiable ray consistency.
\newblock In {\em Proceedings of the IEEE Conference on Computer Vision and
  Pattern Recognition}, pages 2626--2634, 2017.

\bibitem{vaswani2017transformer}
A.~Vaswani, N.~Shazeer, N.~Parmar, J.~Uszkoreit, L.~Jones, A.~N. Gomez,
  {\L}.~Kaiser, and I.~Polosukhin.
\newblock Attention is all you need.
\newblock In {\em Neural Information Processing Systems}, pages 5998--6008,
  2017.

\bibitem{wah2011cub}
C.~Wah, S.~Branson, P.~Welinder, P.~Perona, and S.~Belongie.
\newblock The {Caltech-UCSD Birds-200-2011} dataset.
\newblock Technical Report CNS-TR-2011-001, California Institute of Technology,
  2011.

\bibitem{wang2018pix2pixhd}
T.-C. Wang, M.-Y. Liu, J.-Y. Zhu, A.~Tao, J.~Kautz, and B.~Catanzaro.
\newblock High-resolution image synthesis and semantic manipulation with
  conditional {GANs}.
\newblock In {\em IEEE Conference on Computer Vision and Pattern Recognition
  (CVPR)}, pages 8798--8807, 2018.

\bibitem{BMVC2017_99}
O.~Wiles and A.~Zisserman.
\newblock Silnet : Single- and multi-view reconstruction by learning from
  silhouettes.
\newblock In G.~B. Tae-Kyun~Kim, Stefanos~Zafeiriou and K.~Mikolajczyk,
  editors, {\em Proceedings of the British Machine Vision Conference (BMVC)},
  pages 99.1--99.13. BMVA Press, September 2017.

\bibitem{wu2017marrnet}
J.~Wu, Y.~Wang, T.~Xue, X.~Sun, B.~Freeman, and J.~Tenenbaum.
\newblock Marrnet: 3d shape reconstruction via 2.5d sketches.
\newblock In {\em Neural Information Processing Systems}, pages 540--550, 2017.

\bibitem{wu2016learning}
J.~Wu, C.~Zhang, T.~Xue, B.~Freeman, and J.~Tenenbaum.
\newblock Learning a probabilistic latent space of object shapes via 3d
  generative-adversarial modeling.
\newblock In {\em Advances in Neural Information Processing Systems}, pages
  82--90, 2016.

\bibitem{xiang2014pascal}
Y.~Xiang, R.~Mottaghi, and S.~Savarese.
\newblock Beyond {PASCAL}: A benchmark for {3D} object detection in the wild.
\newblock In {\em IEEE Winter Conference on Applications of Computer Vision
  (WACV)}, 2014.

\bibitem{xie2018learning}
J.~Xie, Z.~Zheng, R.~Gao, W.~Wang, S.-C. Zhu, and Y.~Nian~Wu.
\newblock Learning descriptor networks for 3d shape synthesis and analysis.
\newblock In {\em Proceedings of the IEEE conference on computer vision and
  pattern recognition}, pages 8629--8638, 2018.

\bibitem{xu2018attngan}
T.~Xu, P.~Zhang, Q.~Huang, H.~Zhang, Z.~Gan, X.~Huang, and X.~He.
\newblock {AttnGAN}: Fine-grained text to image generation with attentional
  generative adversarial networks.
\newblock In {\em IEEE Conference on Computer Vision and Pattern Recognition
  (CVPR)}, pages 1316--1324, 2018.

\bibitem{yan2016perspective}
X.~Yan, J.~Yang, E.~Yumer, Y.~Guo, and H.~Lee.
\newblock Perspective transformer nets: Learning single-view 3d object
  reconstruction without 3d supervision.
\newblock In {\em Advances in Neural Information Processing Systems}, pages
  1696--1704, 2016.

\bibitem{yang20173d}
B.~Yang, H.~Wen, S.~Wang, R.~Clark, A.~Markham, and N.~Trigoni.
\newblock 3d object reconstruction from a single depth view with adversarial
  learning.
\newblock In {\em Proceedings of the IEEE International Conference on Computer
  Vision Workshops}, pages 679--688, 2017.

\bibitem{yang2018learning}
G.~Yang, Y.~Cui, S.~Belongie, and B.~Hariharan.
\newblock Learning single-view 3d reconstruction with limited pose supervision.
\newblock In {\em Proceedings of the European Conference on Computer Vision
  (ECCV)}, pages 86--101, 2018.

\bibitem{yang2017lrgan}
J.~Yang, A.~Kannan, D.~Batra, and D.~Parikh.
\newblock {LR-GAN:} layered recursive generative adversarial networks for image
  generation.
\newblock In {\em International Conference on Learning Representations (ICLR)},
  2017.

\bibitem{yaz2019ganaveraging}
Y.~Yaz{\i}c{\i}, C.-S. Foo, S.~Winkler, K.-H. Yap, G.~Piliouras, and
  V.~Chandrasekhar.
\newblock The unusual effectiveness of averaging in {GAN} training.
\newblock In {\em International Conference on Learning Representations (ICLR)},
  2019.

\bibitem{zhang2018sagan}
H.~Zhang, I.~Goodfellow, D.~Metaxas, and A.~Odena.
\newblock Self-attention generative adversarial networks.
\newblock In {\em International Conference on Machine Learning (ICML)}, 2019.

\bibitem{zhang2017stackgan}
H.~Zhang, T.~Xu, H.~Li, S.~Zhang, X.~Wang, X.~Huang, and D.~N. Metaxas.
\newblock {StackGAN}: Text to photo-realistic image synthesis with stacked
  generative adversarial networks.
\newblock In {\em IEEE International Conference on Computer Vision (ICCV)},
  pages 5907--5915, 2017.

\bibitem{zhang2018stackganpp}
H.~Zhang, T.~Xu, H.~Li, S.~Zhang, X.~Wang, X.~Huang, and D.~N. Metaxas.
\newblock {StackGAN}++: Realistic image synthesis with stacked generative
  adversarial networks.
\newblock {\em IEEE Transactions on Pattern Analysis and Machine Intelligence
  (TPAMI)}, 41(8):1947--1962, 2018.

\bibitem{zhang2018unreasonable}
R.~Zhang, P.~Isola, A.~A. Efros, E.~Shechtman, and O.~Wang.
\newblock The unreasonable effectiveness of deep features as a perceptual
  metric.
\newblock In {\em IEEE Conference on Computer Vision and Pattern Recognition
  (CVPR)}, pages 586--595, 2018.

\bibitem{zhao2019imagelayout}
B.~Zhao, L.~Meng, W.~Yin, and L.~Sigal.
\newblock Image generation from layout.
\newblock In {\em IEEE Conference on Computer Vision and Pattern Recognition
  (CVPR)}, pages 8584--8593, 2019.

\bibitem{zhu2017cyclegan}
J.-Y. Zhu, T.~Park, P.~Isola, and A.~A. Efros.
\newblock Unpaired image-to-image translation using cycle-consistent
  adversarial networks.
\newblock In {\em IEEE International Conference on Computer Vision (ICCV)},
  pages 2223--2232, 2017.

\bibitem{zhu2018visual}
J.-Y. Zhu, Z.~Zhang, C.~Zhang, J.~Wu, A.~Torralba, J.~Tenenbaum, and
  B.~Freeman.
\newblock Visual object networks: Image generation with disentangled 3d
  representations.
\newblock In {\em Neural Information Processing Systems}, pages 118--129, 2018.

\bibitem{zhu2017rethinking}
R.~Zhu, H.~Kiani~Galoogahi, C.~Wang, and S.~Lucey.
\newblock Rethinking reprojection: Closing the loop for pose-aware shape
  reconstruction from a single image.
\newblock In {\em Proceedings of the IEEE International Conference on Computer
  Vision}, pages 57--65, 2017.

\end{thebibliography}
}

\clearpage
\newpage
\appendix

\section{Supplementary material}
\subsection{Detailed architecture}
\label{sec:appendix-architecture}

\begin{figure}[h]
  \centering
  \includegraphics[width=\textwidth]{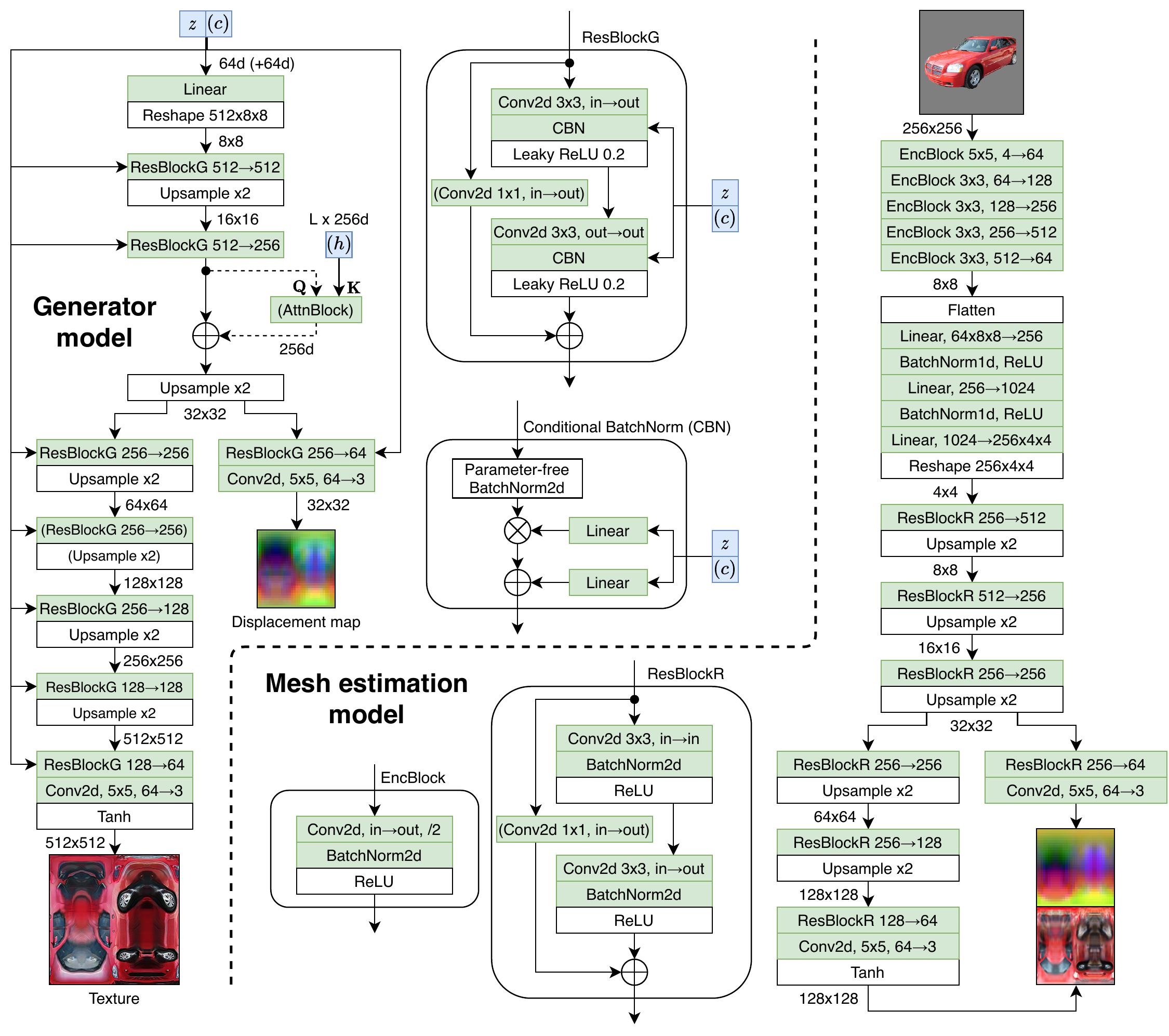}
  \vspace{-7mm}
  \caption{Generator architecture (left) and mesh estimation model (right). Green blocks comprise learnable parameters, whereas white ones are parameter-free. Dashed lines and blocks in parentheses represent optional connections which depend on the specific setting. We indicate the feature map resolution in a given position next to arrows (e.g. 128$\times$128). $512\rightarrow256$ denotes ``512 input channels, 256 output channels''. ``/2'' in convolutional layers denotes ``stride 2''; it is one if not indicated. }
  \label{fig:appendix-generator-architecture}
\end{figure}

\paragraph{Generator.} \autoref{fig:appendix-generator-architecture} (left) shows the detailed architecture of the generator of our GAN. As mentioned in \autoref{sec:implementation-details}, the random vector $\mathbf{z}$ is fed to every conditional batch normalization (CBN) layer as well as to the input layer, which matches the strategy adopted by many state-of-the-art GANs for 2D image generation \cite{miyato2018cgans, zhang2018sagan, brock2018biggan}. A CBN layer consists of a parameter-free batch normalization (i.e. without a learned affine transformation) followed by a gain $\boldsymbol{\gamma}$ and bias $\boldsymbol{\beta}$ conditioned upon $\mathbf{z}$ via a learned linear layer. In settings conditioned on class labels, we concatenate a learned embedding $\mathbf{c}$ to $\mathbf{z}$, which is shared among all layers. The network follows a ResNet architecture where feature maps are progressively upsampled using nearest-neighbor interpolation after each residual block \texttt{ResBlockG}. This block consists of two convolutional layers wrapped by a skip-connection. If the number of input channels differs from the number of output channels, the skip-connection is learned. To accommodate for the varying output resolutions for mesh and texture, the generator branches out at 32$\times$32 resolution. While the figure shows the architecture for 512$\times$512 textures, to generate textures at 256$\times$256 we simply remove one 256$\rightarrow$256 \texttt{ResBlockG} block. For presentation purposes, we report square resolutions (e.g. 512$\times$512), but in practice we only need to generate half of the feature map (e.g. 256$\times$512) since we enforce symmetry across the $x$ axis as mentioned in \autoref{sec:implementation-details}. The output textures and displacement maps are then simply padded with their reflection. On the other hand, the discriminator always observes full textures as \emph{pseudo-}ground-truth textures are asymmetric.

\paragraph{Attention mechanism.} If the model is conditioned on text using an attention mechanism, we add an attention block right before the texture/mesh branch, so that the module influences both mesh and texture. We adopt a dot-product formulation similar to \cite{xu2018attngan}, in which the attention weights are computed as $\mathrm{softmax}(\mathbf{Q}\mathbf{K}^T)$. The queries $\mathbf{Q}$ correspond to the flattened convolutional feature map from the generator, and the keys $\mathbf{K}$ are obtained by passing each RNN hidden state $\mathbf{h}_l$ ($l \in \{1 .. L\}$, where $L$ is the sentence length) through a learned linear layer. The RNN is the pretrained bidirectional LSTM encoder from \cite{xu2018attngan}, and hidden states are 256-dimensional.

\paragraph{Mesh estimation model.} \autoref{fig:appendix-generator-architecture} (right) shows the detailed architecture of the mesh estimation model that we use for differentiable rendering (the first step of our algorithm as described in \autoref{fig:step1}). The natural image is concatenated to the segmentation mask (3+1 channels) and fed to a convolutional encoder. The representation is then flattened to a dense representation and passed through a series of linear layers. Finally, it is passed through a ResNet decoder whose architecture resembles that of the GAN generator. Since we are not interested in producing high-quality textures in this step as they are discarded, the texture resolution in this model is only 128$\times$128, which results in faster training.

\begin{figure}[h]
  \centering
  \includegraphics[width=\textwidth]{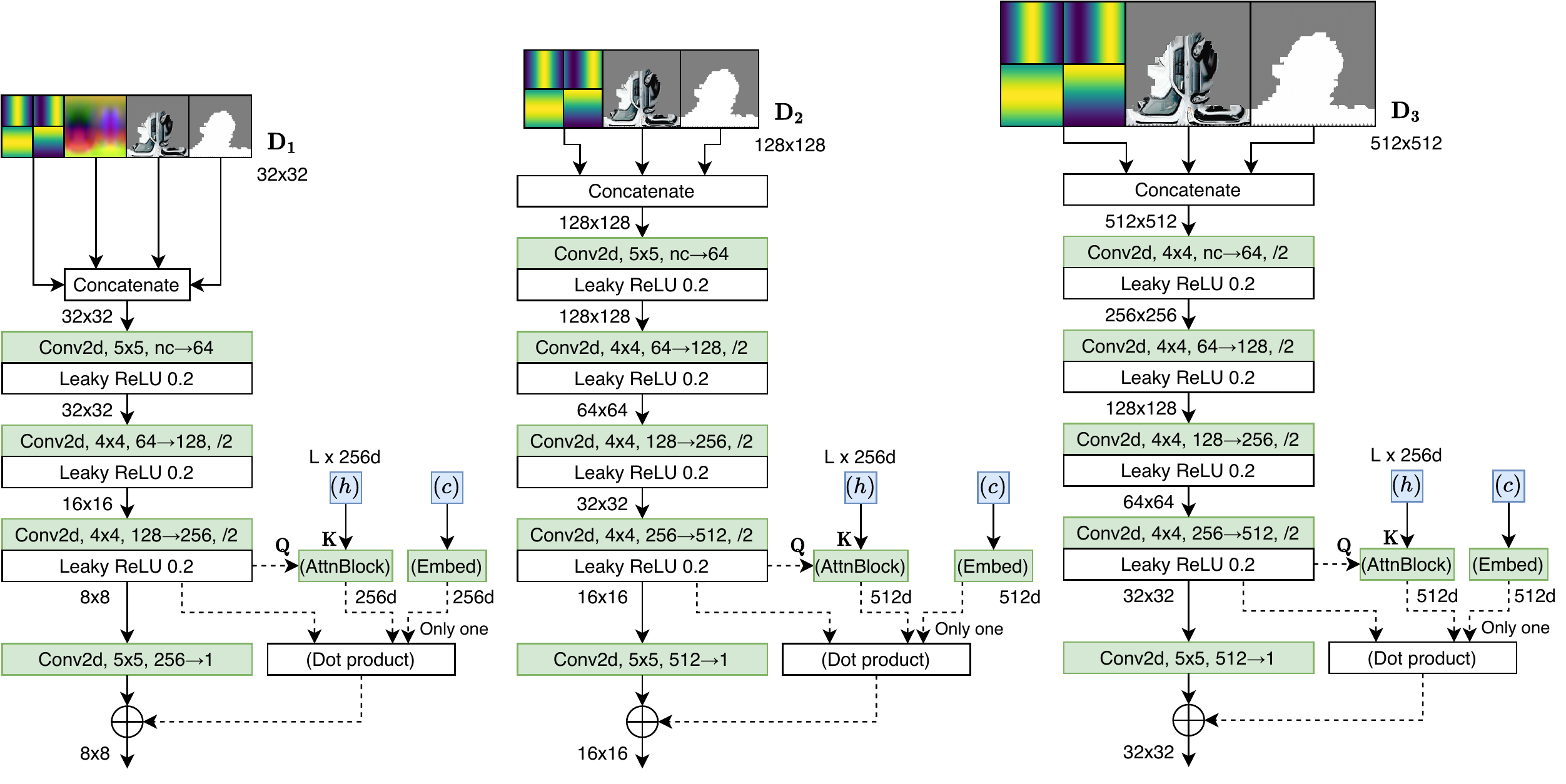}
  \vspace{-7mm}
  \caption{Multi-scale discriminator architecture for our biggest model (512$\times$512 texture resolution). Only $\mathbf{D_1}$ discriminates the mesh, while $\mathbf{D_2}$ and $\mathbf{D_3}$ are texture discriminators. Dashed lines describe the optional connections for \emph{projection discrimination} \cite{miyato2018cgans} in conditional settings, where feature maps are combined \emph{either} with a learned embedding (for settings conditioned on classes or attributes) or with the values of an attention block (for settings conditioned on text), not both.}
  \label{fig:appendix-discriminator-architecture}
\end{figure}

\paragraph{Discriminator.} The architecture of our multi-scale discriminator is depicted in \autoref{fig:appendix-discriminator-architecture}. In the most complex setting (used by the CUB model at 512$\times$512), textures are discriminated at three scales: 32$\times$32 ($\mathbf{D_1}$), 128$\times$128 ($\mathbf{D_2}$), and 512$\times$512 ($\mathbf{D_3}$). The smallest discriminator $\mathbf{D_1}$ is also a mesh discriminator. Following the general strategy of patch-based discriminators \cite{isola2017pix2pix}, our discriminators are relatively simple as they only consist of a series of spectrally-normalized convolutional layers.
GANs have been shown to produce checkerboard artifacts \cite{odena2016deconvolution} depending on the choice of kernel sizes and strides in the discriminator. While humans do not perceive these to be particularly severe in images, checkerboard artifacts in the \emph{displacement map} must be avoided as they might lead to noticeable mesh distortions. The generator already uses upsampling instead of transposed convolutions (which mitigates this issue), but we also carefully design the discriminator such that the kernel size of convolutions is divisible by the stride, ensuring that the gradient norms are uniform across pixels (see \cite{odena2016deconvolution} for further details). To this end, we use 5$\times$5 convolutions in layers with stride 1, and 4$\times$4 convolutions in layers with stride 2. $\mathbf{D_1}$ consists of 4 layers and has a relatively small receptive field. We explored a varying number of layers and different strides, but they always led to worse results. $\mathbf{D_2}$ and $\mathbf{D_3}$ consist of 5 layers and are identical except for the stride of the first layer, which is 1 for $\mathbf{D_2}$ and 2 for $\mathbf{D_3}$. In the experiments with a texture resolution of 256$\times$256, we only use $\mathbf{D_1}$ and $\mathbf{D_2}$, where the latter directly discriminates at 256$\times$256. For Pascal3D+ at 512$\times$512, we found a small empirical advantage in dropping $\mathbf{D_2}$ and doubling the weight of $\mathbf{D_3}$'s loss.
Finally, to incorporate conditional information, we use \emph{projection discrimination}  \cite{miyato2018cgans}, %
in which we compute a pixel-wise dot product between the last feature map and a learned class embedding (a vector), and add it to the output. If the model is conditioned on text, we replace the class embedding with the output of an attention block. Each discriminator learns its own set of weights for the embeddings or the attention block.\looseness=-1

\paragraph{Model complexity.} \autoref{tbl:appendix-complexity} summarizes the complexity of our models in different settings, expressed as the number of learnable parameters. Since the semantic alignment of our pose-independent representation facilitates the modeling task, our approach can successfully work with relatively small models ($\approx$10M generator parameters). We also found it beneficial to adopt simple discriminators compared to the generator (which is more powerful).

\begin{table}[ht]
\caption{Number of learnable parameters for different variants of our model.}
\label{tbl:appendix-complexity}
\centering
\renewcommand{\arraystretch}{1.15}
\resizebox{0.72\textwidth}{!}{
\begin{tabular}{l|l|l|l|l|l|l|l}
Model                                  & Resolution & $\mathbf{G}$      & $\mathbf{D_1}$                     & $\mathbf{D_2}$                     & $\mathbf{D_3}$    & $\mathbf{D}$ Total & RNN   \\ \hline\hline
\multirow{2}{*}{Unconditional} & 512x512    & 11.58M & \multirow{2}{*}{0.68M} & \multirow{2}{*}{2.77M} & 2.77M & 6.22M   & -     \\ \cline{2-3} \cline{6-8} 
                                       & 256x256    & 10.33M &                        &                        & -     & 3.45M   & -     \\ \hline\hline
\multirow{2}{*}{CUB conditional}       & 512x512    & 13.06M & \multirow{2}{*}{0.73M} & \multirow{2}{*}{2.88M} & 2.88M & 6.49M   & -     \\ \cline{2-3} \cline{6-8} 
                                       & 256x256    & 11.75M &                        &                        & -     & 3.61M   & -     \\ \hline\hline
CUB text                               & 512x512    & 11.64M & 0.75M                  & 2.91M                  & 2.91M & 6.57M   & 2.08M \\ \hline\hline
\multirow{2}{*}{P3D conditional}       & 512x512    & 13.05M & \multirow{2}{*}{0.69M} & \multirow{2}{*}{2.79M} & 2.79M & 6.27M   & -     \\ \cline{2-3} \cline{6-8} 
                                       & 256x256    & 11.74M &                        &                        & -     & 3.48M   & -    
\end{tabular}
}
\end{table}

\subsection{Additional implementation details}
\label{sec:appendix-implementation-details}

\paragraph{Mesh estimation.} Since our \emph{convolutional mesh} representation already encourages meshes to be smooth, our reconstruction model requires less regularization than similar frameworks based on fully-connected networks. We only found it beneficial to regularize the model with a smoothness loss $\mathcal{L}_\text{flat}$ \cite{kato2018n3mr} at a very low strength $\alpha = 0.00005$, and no Laplacian regularization \cite{sorkine2004laplacian} (unlike \cite{kato2018n3mr, kanazawa2018cmr, chen2019dibr} which all use this form of regularization). $\mathcal{L}_\text{flat}$ encourages the normals of neighboring faces to have similar directions, and is defined as follows:
\setlength{\abovedisplayskip}{1mm}
\setlength{\belowdisplayskip}{1mm}
\begin{align*}
    \mathcal{L}_\text{flat} = \alpha \frac{1}{|E|} \sum_{i, j \in E} (1 - \cos\theta_{ij})^2
\end{align*}
where $E$ is the set of all edges and $\cos\theta_{ij}$ is the cosine similarity between the normals of the faces $i$ and $j$. In practice this is implemented by computing the dot product between the two normals.

We additionally observe that the initialization strategy of this model as well as early training iterations have a significant impact on the final result. Bad configurations such as self-intersecting meshes or vertices outside the camera frustum can cause the model to get stuck in bad local minima from which it cannot easily recover. %
This is especially the case for typical Gaussian initialization schemes in neural networks, which cause the mesh to start in an already self-intersected state for a spherical mesh template with radius 1 (our case). To ensure convergence and generate smooth meshes without self-intersections, we found it helpful to \emph{(i)} zero-initialize the final layer of the mesh branch, which ensures that the first iteration starts with a smooth sphere, and \emph{(ii)} adopt a warm-up phase where $\mathcal{L}_\text{flat}$ starts at a moderate strength $\alpha = 0.0005$ and linearly decays for 100 iterations, settling at the low-strength value mentioned above. In the GAN generator we also zero-initialize the final layer of the mesh head, but $\mathcal{L}_\text{flat}$ only uses a fixed $\alpha = 0.0001$ and no warm-up.

In \autoref{sec:method}, we mention that our camera projection model is a weak-perspective model. This model is a good approximation for photographs shot with high levels of zoom or that depict small objects, which is the case for birds (CUB dataset). However, we observed that the weak-perspective assumption is not a good fit for Pascal3D+, since most images are shot from a close range and present a significant degree of perspective distortion due to cars having elongated shapes. Therefore, for Pascal3D+ we augment the camera model with a learnable perspective correction term $z_0$, without however advancing to a full perspective model as we do not have enough information. $z_0$ is a scalar that describes the distance from the camera to the center of the object, and assumes that the object is centered. The $x, y$ coordinates of each vertex in camera space are then multiplied with a factor $(z_0 + z/2)/(z_0 - z/2)$, where $z$ is the depth of the vertex. Note that, as $z_0$ approaches infinity, the factor approaches 1 and the camera model reverts back to a weak-perspective model. This term is learned for every image in the dataset and is parameterized as $z_0 = 1 + e^w$ ($w$ is a learnable parameter), which ensures \emph{(i)} positivity, and \emph{(ii)} that the transformed vertices lie inside the camera frame. While this aspect is not central to our approach, we found it helpful as it can slightly improve qualitative results even with approximate estimates.\looseness=-1

\paragraph{Construction of the pose-independent representation.} In \autoref{sec:method} we mention that we use the gradient from the differentiable renderer to produce the UV visibility mask which is used for masking projected textures. In practice, deep learning frameworks do not compute full Jacobians but only gradients of scalars (i.e.\ Jacobian-vector multiplication). However, the Jacobian of the rendering operation w.r.t. the texture has a structure such that it is zero for all texels that are not visible in the rendered image (i.e.\ are occluded) and non-zero elsewhere\footnote{This property holds for DIB-R \cite{chen2019dibr} but may not hold for all differentiable renderers.}. Based on this observation, it suffices to compute the average or sum of rendered pixels to reduce the image to a scalar which can then be differentiated with respect to a dummy texture. The same result can also be achieved by computing a Jacobian-vector multiplication with a vector of ones, which is what we do in our implementation.

\begin{figure}[t]
  \centering
  \includegraphics[width=\textwidth]{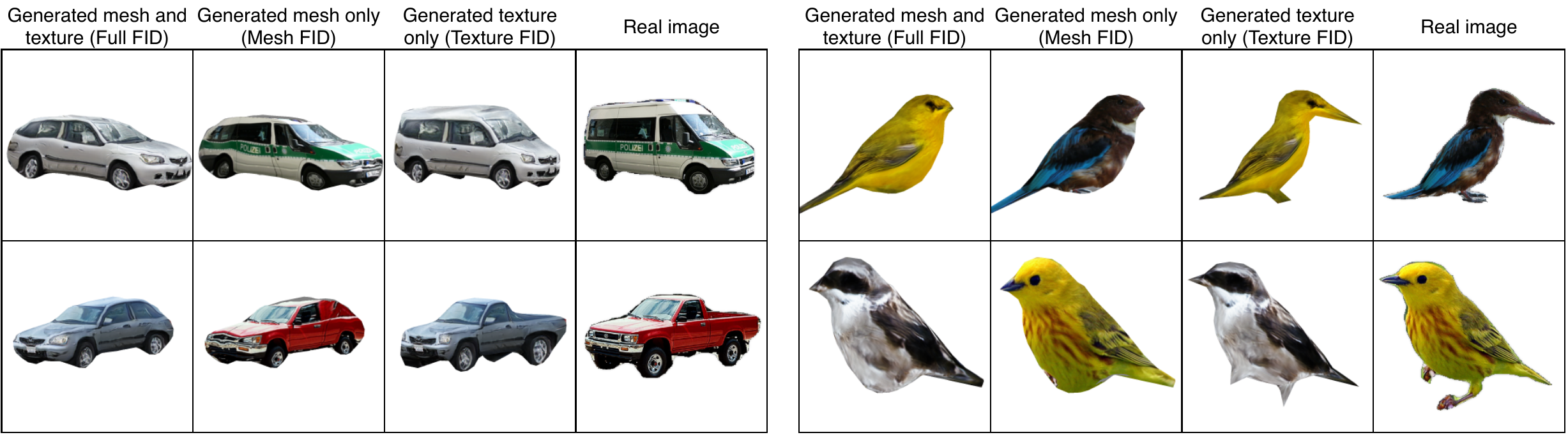}
  \vspace{-5mm}
  \caption{Examples of images on which we compute FID scores. Images are rendered from the viewpoint corresponding to \emph{Real image} (a randomly-selected image from the training set). In the \emph{Mesh FID} scenario, we render the generated mesh using the \emph{pseudo-}ground-truth texture from the real image. In the \emph{Texture FID}, the ``real'' mesh is textured using the generated texture. In the \emph{Full FID} and \emph{Mesh FID} of the top-left van we can observe that the silhouette of the mesh looks fine but straight lines and stripes present a ``wobbling'' effect caused by the underlying mesh, while in the \emph{Texture FID} (which does not use generated meshes) the lines appear more straight.}
  \label{fig:appendix-fid}
\end{figure}

\paragraph{FID evaluation.} To give more context to \autoref{sec:evaluation}, where we introduce our evaluation methodology, in \autoref{fig:appendix-fid} we show some actual examples of rendered images on which we compute FID scores. The \emph{Full FID} (our main metric) is computed on generated meshes coupled with generated textures, and evaluates the generation quality as a whole. However, it is also interesting to propose variations of this metric that can evaluate mesh and texture quality separately. Therefore, in the \emph{Mesh FID} we use the \emph{pseudo-}ground-truth texture from the image corresponding to the random viewpoint we choose for rendering, which makes the evaluation independent of generated textures. Likewise, in the \emph{Texture FID} we use meshes estimated using the differentiable renderer instead of the ones generated by our GAN. In all experiments, we generate as many images as there are in the set we compare to, since the FID is sensitive to the number of generated images.
Finally, to evaluate text conditioning on CUB, we sample a random caption (out of 10 captions) for each image we generate.

\paragraph{Pascal3D+ annotations.} To demonstrate conditional generation on P3D, we collected shape and color annotations for the ImageNet subset of this dataset (i.e. the one we use to train our GAN). Although ImageNet images are already identified by their \emph{synsets}, we found these to be unreliable and opted instead for collecting our own annotations. The set of labels and corresponding frequencies are summarized in \autoref{tbl:appendix-p3d-data}. For consistency, all labels were collected by one annotator. Some categories (e.g. \emph{F1}, \emph{convertible}, and \emph{oldtimer}) comprise a very low number of samples, which leads to unsatisfactory results on these classes in conditional settings. Nonetheless, this issue can be mitigated by collecting more data. Finally, although the ImageNet subset consists of $\approx$5.5k images, only 4.7k are usable as some are filtered out by the structure-from-motion routine of \cite{kanazawa2018cmr} due to unreliable pose estimates.\looseness=-1

\begin{table}[ht]
\caption{Relevant statistics for the P3D annotations we collected.}
\label{tbl:appendix-p3d-data}
\centering
\renewcommand{\arraystretch}{1.15}
\resizebox{\textwidth}{!}{
\begin{tabular}{l|lllllllllll|l}
Class & Sedan & Hatchback & SUV & Station wagon & Van  & Pickup & Coupé  & City   & F1    & Convertible & Oldtimer & Total \\
\# images   & 1137  & 851       & 814 & 691            & 674  & 649    & 295    & 193    & 119   & 39          & 13       & 5475  \\ \hline\hline
Color & Gray  & Black     & Red & White          & Blue & Green  & Yellow & Orange & Brown & Purple      & Pink     & Total \\
\# images   & 1534  & 863       & 833 & 832            & 697  & 252    & 231    & 126    & 52    & 30          & 25       & 5475 
\end{tabular}
}
\end{table}

\subsection{Additional results}
\label{sec:appendix-results}

\begin{figure}[t]
  \centering
  \includegraphics[width=\textwidth]{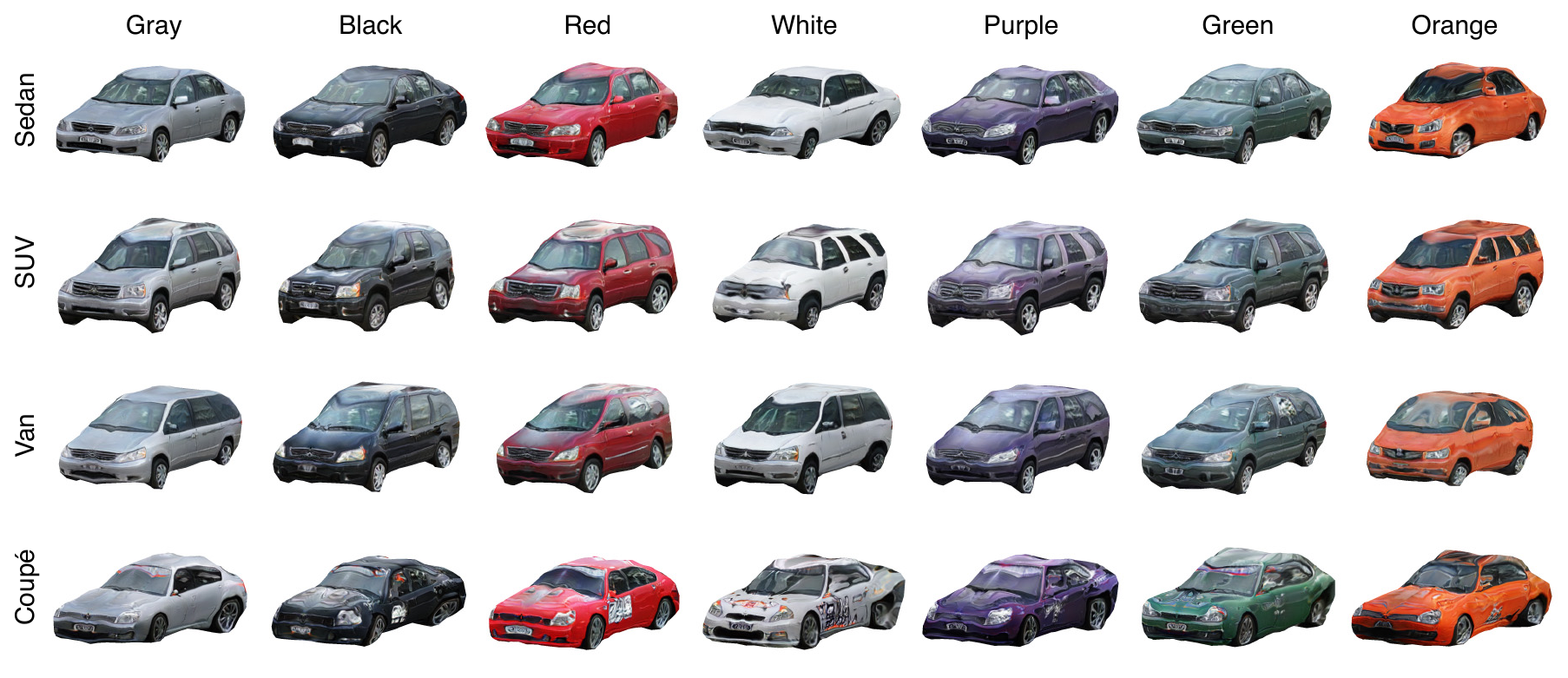}
  \vspace{-6mm}
  \caption{Generation on P3D with one varying factor at a time (color and shape) and a fixed random vector $\mathbf{z}$. As can be seen, representations are relatively disentangled. In the bottom row, the class \emph{coupé} is often associated with race cars, which causes stickers to appear on the body.}
  \label{fig:appendix-disentanglement}
\end{figure}

\paragraph{Disentanglement.} Compared to a generative model for 2D images, a 3D generative model naturally disentangles pose and appearance. Furthermore, the use of \emph{triangle meshes} with UV-mapped textures ensures that shape is disentangled from color, essentially allowing for texture transfer between meshes (we use this property for our \emph{Texture FID} and \emph{Mesh FID} evaluation, as shown in \autoref{fig:appendix-fid}). Another interesting observation is that conditional models enable further disentanglement of aspects of variation. For instance, on P3D we can control shape and color separately as shown in \autoref{fig:appendix-disentanglement}, and the latent space is structured enough to allow for interpolation of these aspects (\autoref{fig:interpolation}). In this setting, the random vector $\mathbf{z}$ can be used to control the style of the object.

\begin{figure}[tp]
  \centering
  \vspace{-6mm}
  \includegraphics[width=\textwidth]{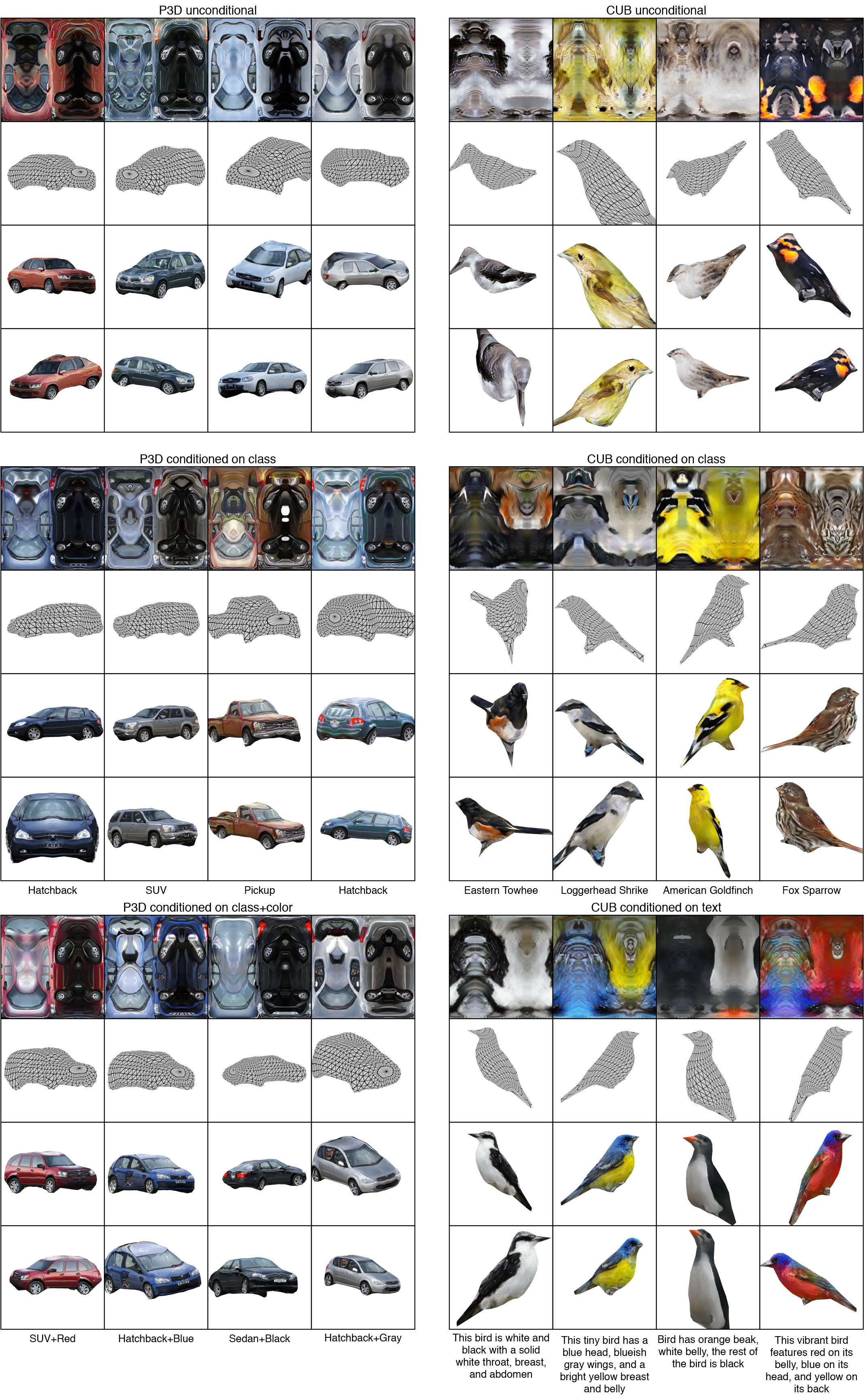}
  \vspace{-7mm}
  \caption{Qualitative results for all settings, on both P3D (left) and CUB (right). First row = texture; second row = wireframe mesh; third and fourth rows = textured object from two random views.}
  \label{fig:appendix-qualitative}
\end{figure}

\paragraph{Additional qualitative results.} In \autoref{fig:appendix-qualitative}, we show additional qualitative results grouped by type of conditioning. Our approach successfully generates meshes in both conditional and unconditional settings. In the figure, we additionally show untextured (i.e. \emph{wireframe}) meshes, which highlights the smoothness of our convolutional mesh representation.

\paragraph{Demo video.} The supplementary material includes a video where we show more results, including latent space interpolation, disentangled generation, generation from text, and visualization of the attention mechanism on models conditioned on text.

\end{document}